\documentclass[sigconf]{acmart}
\usepackage{times}
\usepackage{amsmath}
\usepackage{booktabs}
\usepackage{graphicx}
\usepackage{subfigure}
\usepackage{footnote}
\usepackage{color}
\usepackage{epstopdf}
\usepackage{hyperref}

\usepackage{url}
\usepackage{float}
\usepackage{microtype}
\usepackage{makecell}
\usepackage{multirow}
\usepackage{hyperref}
\usepackage{listings}
\usepackage{color}
\definecolor{mygreen}{rgb}{0,0.6,0}
\definecolor{mygray}{rgb}{0.5,0.5,0.5}
\definecolor{mymauve}{rgb}{0.58,0,0.82}

\usepackage[linesnumbered, ruled, boxed]{algorithm2e}

\newcommand{\boldpara}[1]{\vspace{0.12cm}\noindent \textbf{#1}}

\newcommand\BibTeX{B\textsc{ib}\TeX}

\AtBeginDocument{%
  \providecommand\BibTeX{{%
    \normalfont B\kern-0.5em{\scshape i\kern-0.25em b}\kern-0.8em\TeX}}}

\setcopyright{acmcopyright}
\copyrightyear{2021}
\acmYear{2021}
\setcopyright{acmcopyright}\acmConference[CIKM '21]{Proceedings of the 30th ACM International Conference on Information and Knowledge Management}{November 1--5, 2021}{Virtual Event, QLD, Australia}
\acmBooktitle{Proceedings of the 30th ACM International Conference on Information and Knowledge Management (CIKM '21), November 1--5, 2021, Virtual Event, QLD, Australia}
\acmPrice{15.00}
\acmDOI{10.1145/3459637.3481911}
\acmISBN{978-1-4503-8446-9/21/11}

\title{EasyTransfer: A Simple and Scalable Deep Transfer Learning Platform for NLP Applications}

\newcommand\blfootnote[1]{%
  \begingroup
  \renewcommand\thefootnote{}\footnote{#1}%
  \addtocounter{footnote}{-1}%
  \endgroup
}

\begin{document}
\fancyhead{}

\author{Minghui Qiu$^1$, Peng Li$^{1*}$, Chengyu Wang$^{1*}$, Haojie Pan$^{1*}$, Ang Wang$^1$, Cen Chen$^1$, Xianyan Jia$^1$, Yaliang Li$^1$, Jun Huang$^1$, Deng Cai$^2$, Wei Lin$^1$}
\affiliation{$^1$ Alibaba Group $^2$ State Key Lab of CAD\&CG, Zhejiang University\country{China}}

\renewcommand{\shortauthors}{}
\renewcommand{\authors}{M. Qiu, P. Li, C. Wang, H. Pan, A. Wang, C. Chen, X. Jia, Y. Li, J. Huang, D. Cai, W. Lin}

\email{{minghui.qmh,jerry.lp,chengyu.wcy,haojie.phj,chencen.cc,wangang.wa,xianyan.xianyanjia}@alibaba-inc.com}
\email{{yaliang.li,huangjun.hj,weilin.lw}@alibaba-inc.com, dengcai@cad.zju.edu.cn}



\begin{abstract}
The literature has witnessed the success of leveraging Pre-trained Language Models (PLMs) and Transfer Learning (TL) algorithms to a wide range of Natural Language Processing (NLP) applications, yet it is not easy to build an easy-to-use and scalable TL toolkit for this purpose. To bridge this gap, the EasyTransfer platform is designed to develop deep TL algorithms for NLP applications.
EasyTransfer is backended with a high-performance and scalable engine for efficient training and inference, and also integrates comprehensive deep TL algorithms, to make the development of industrial-scale TL applications easier. In EasyTransfer, the built-in data and model parallelism strategies, combined with AI compiler optimization, show to be 4.0x faster than the community version of distributed training. 
EasyTransfer supports various NLP models in the ModelZoo, including mainstream PLMs and multi-modality models. It also features various in-house developed TL algorithms, together with the AppZoo for NLP applications. 
The toolkit is convenient for users to quickly start model training, evaluation, and online deployment. 
EasyTransfer is currently deployed at Alibaba to support a variety of business scenarios, including item recommendation, personalized search, conversational question answering, etc. Extensive experiments on real-world datasets and online applications show that EasyTransfer is suitable for online production with cutting-edge performance for various applications.\blfootnote{$^{*}$~Equal contributions.} The source code of EasyTransfer is released at Github~\footnote{https://github.com/alibaba/EasyTransfer}.
\end{abstract}


\maketitle

\section{Introduction}
Transfer Learning (TL) is a rapidly growing field of machine learning that aims to improve the learning of a data-deficient task by transferring knowledge from related data-sufficient tasks~\cite{pan2009survey,tan2018survey,DBLP:journals/corr/abs-1911-02685}. 
Witnessing the great representation learning abilities of deep neural networks, neural TL methods, i.e., deep transfer learning, have gained increasing popularity and are shown to be effective for a wide variety of applications~\cite{mou:EMNLP2016,Zhilin,TSHS,TLMRCNN}.
Despite the success of deep TL methods for real-world applications, it is still very challenging to design a deep TL framework. There are a few earliest attempts to build TL toolkits. Notable projects include: 
\begin{itemize}
    \item The NVIDIA Transfer Learning Toolkit (TLT)~\footnote{https://developer.nvidia.com/transfer-learning-toolkit} is a python-based toolkit for training models and customizing them with users' own datasets. It mainly focuses on computer vision.
    \item Amazon Xfer~\footnote{https://github.com/amzn/xfer} is an MXNet library that largely automates deep TL. It contains the ``ModelHandler'' component to extract features from pre-trained models and the ``Repurposer'' component to re-purpose models for target tasks.
    \item Transfer Learning Toolkit ~\footnote{https://github.com/FuzhenZhuang/Transfer-Learning-Toolkit} 
    integrates five types of models, namely feature-based, concept-based, parameter-based, instance-based, and deep-learning-based. 
    \item The Huggingface Transformers toolkit~\footnote{https://github.com/huggingface/transformers} specifically addresses model-finetuning, especially for BERT-like models. It is backended by PyTorch and Tensorflow 2.0 and integrates Pre-trained Language Models (PLMs).
\end{itemize}

\begin{table*}[th!]
    \centering
    \caption{An overview of transfer learning toolkits.}
    \begin{small}
    \begin{tabular}{l | c c c c c c c c }
        \toprule
        \multirow{2}{*}{Framework}& \multirow{2}{*}{ModelZoo} & \multirow{2}{*}{AppZoo} & \multicolumn{5}{c}{TL Algorithms} & Production- \\\cline{4-8}
        & & & Fine-tuning & Feature-based & Instance-based & Model-based & Meta Learning & ready\\
        \midrule
        NVIDIA TLT  & \checkmark & & \checkmark & & & & & \\
        Amazon Xfer & \checkmark & & \checkmark & & & & & \\
        Hugginngface Transformers & \checkmark & & \checkmark & &  & \checkmark & & \\
        TL Toolkit& \checkmark & & \checkmark & \checkmark & \checkmark & \checkmark & & \\
        EasyTransfer & \checkmark & \checkmark & \checkmark & \checkmark & \checkmark & \checkmark & \checkmark & \checkmark\\
        \bottomrule
    \end{tabular}
    \end{small}
    \label{tab:overview}
\end{table*}

\boldpara{Challenges.} 
However, when it comes to industrial-scale real-world applications, the above-mentioned toolkits might be less ideal. The reasons are threefold. i) PLMs such as BERT \cite{devlin2019bert}, T5 \cite{raffel2020exploring} and GPT \cite{radford2018improving} are becoming larger and usually have parameters on the billion scale. 
To elaborate, T5-large~\cite{raffel2020exploring} and GPT-3 are with 11 and 175 billion parameters respectively, which exceed the memory limit of modern processors. It is needed to build a toolkit to support scalable distributed training strategies to train and deploy such models efficiently and effectively in real-world applications.
ii) Real-world applications with diverse data properties require different types of TL algorithms, yet there are no TL toolkits available for users to examine rich types of state-of-the-art TL algorithms. There is an increasing need for a toolkit that supports a comprehensive suite of TL algorithms. 
iii) A gap still exists between developing an algorithm for a specific task and deploying the algorithm for online production. For many online applications, it is still a non-trivial task to provide a reliable service with high QPS (Queries Per Second) requirements. In a nutshell, it is highly necessary to develop a comprehensive, industry-scale deep TL toolkit. 

\boldpara{Our Work.} 
In light of these challenges, we develop the EasyTransfer toolkit and release it to the open-source community. EasyTransfer is built with highly scalable distributed training strategies, facilitating super large-scale model training. It supports a comprehensive suite of TL algorithms that can be used in various NLP tasks, providing a unified framework of model training, inference, and deployment for real-world applications. Currently, we have integrated EasyTransfer into a number of deep learning products in Alibaba and observed notable performance gains. Table~\ref{tab:overview} presents an overview of 
existing TL toolkits. Comparing to other toolkits, EasyTransfer provides more comprehensive and scalable TL algorithms and functionalities for developing and deploying NLP applications.

In summary, this paper makes the following contributions:
\begin{itemize}
    \item We are the first to propose a simple and scalable deep TL platform named EasyTransfer to make it easy to develop deep TL algorithms for NLP applications. It is built with efficient distributed training strategies and AI compiler optimization for training super large-scale NLP models;

    \item EasyTransfer provides a rich family of TL algorithms to cover a wide range of applications in industry, including state-of-the-art and in-house developed algorithms where all these algorithms are shown to be effective in real-world applications;
    
    \item EasyTransfer is equipped with ModelZoo, containing more than 20 mainstream PLMs and a multi-modality model. EasyTransfer further integrates AppZoo and simple user interfaces to support for building NLP applications 
    
    \item Extensive experiments have demonstrated the efficiency and scalability of the toolkit, and also the effectiveness in real-world applications. The toolkit has been deployed at Alibaba to support a variety of business scenarios (more than 20+ applications).
    EasyTransfer is seamlessly integrated with Platform of AI (PAI) products~\footnote{https://www.alibabacloud.com/product/machine-learning}, to make it easy for external users outside of Alibaba to conduct model training, evaluation and online deployment on the cloud. 
\end{itemize}

EasyTransfer is released under the Apache 2.0 License and open-sourced at GitHub. The detailed documentation and tutorials are available on this link~\footnote{\url{https://www.yuque.com/easytransfer/cn}}.
The rest of this paper is organized as follows. Section \ref{sec:related} introduces the related work. Section \ref{sec:infra} and Section \ref{sec:algos} present EasyTransfer infrastructure and algorithms.
Extensive experiments in Section~\ref{sec:exp} examine the efficiency, scalability and effectiveness of EasyTransfer.
The last section draws the conclusion.

\section{Related Work}
\label{sec:related}
In this section, we summarize the related work on transformer-based PLMs and various deep TL methods.

\subsection{Transformer-based PLMs}
\boldpara{Transformers.}
For PLMs, transformers can primarily be applied in three ways: encoder-only, decoder-only, and encoder-decoder. The original transformer~\cite{vaswani2017attention} was designed as a machine translation architecture.
BART~\cite{lewis2019bart} and T5~\cite{raffel2020exploring} leverage this architecture to achieve state-of-the-art performance on downstream classification tasks. Previous work leveraging transformers for language modeling such as BERT~\cite{devlin2019bert},  RoBERTa~\cite{liu2019roberta}and ALBERT~\cite{lan2020albert} only use the encoder architecture.
GPT-2~\cite{radford2019language} only uses the decoder architecture to solve text generation tasks. Recent studies also find transformer architectures are effective for modeling cross-modality datasets, e.g. ViLBERT~\cite{lu2019vilbert}, Unicoder-VL~\cite{li2019unicodervl} and VLBERT~\cite{su2020vlbert} for image and text data. EasyTransfer features FashionBERT~\cite{fashionbert}, which studies cross-modality learning with transformers for e-commerce. 

\boldpara{Sparse Transformers.}
The complexity of self-attention in transformers grows quadratically with the sequence length.
Recent studies consider sparse transformers to alleviate this problem, which mainly fall into three groups: pattern-based, memory-based and low-rank. Pattern-based methods blockify attention matrices by limiting the field of view. BlockBERT~\cite{qiu2020blockwise} chunks input sequences into fixed blocks. Sparse Transformer~\cite{child2019generating} combines strided and local attention by assigning half of its heads to the pattern. Memory-based methods such as Longformer~\cite{beltagy2020longformer} use global memory to access the entire sequence. Set Transformers~\cite{lee2019set} applies inducing points from sparse Gaussian process to reduce the complexity. 
Low-rank approximation seeks to build a low-rank representation of full-attention matrices. Linformer~\cite{wang2020linformer} projects the dimension length of keys and values to a lower dimension. EasyTransfer develops a novel sparse attention kernel that reduces memory footprints
while preserving similar downstream task performance.

\subsection{Deep TL Methods}
In the literature, different types of TL models have been proposed over the years, depending on how the knowledge is shared~\cite{pan2009survey,tan2018survey}. In this work, we categorize deep TL models into five types, i.e., \textit{model fine-tuning}, \textit{feature-based}, \textit{instance-based}, \textit{model-based}, and \textit{meta learning}, detailed in the following. 

\boldpara{Model fine-tuning.} It is a simple but effective TL paradigm, where the pre-trained model in the source domain is used as the initialization for continued training on the target domain data. Model fine-tuning has been broadly used to reduce the number of labeled data needed for learning new tasks and tasks in new domains~\cite{he2016deep,howard2018universal,devlin2019bert}. 

\boldpara{Feature-based methods.} This type of methods aims to locate a common feature space that can reduce the differences between the source and target domains. 
They transform features from one domain to be closer to another, or project features of different domains into a common latent space where the feature distributions are close to each other~\cite{shen2017wasserstein}. 
With the recent advances of deep learning, different NN-based TL frameworks have been proposed for feature-based TL~\cite{shen2017wasserstein,mou:acl2016,mou:EMNLP2016,liu2017adversarial,drss}. 
A simple but widely used framework is to train a shared NN to learn a shared feature space~\cite{mou:EMNLP2016}. Another representative framework is to employ a shared NN and two domain-specific NNs to respectively derive a shared feature space and two domain-specific feature spaces ~\cite{liu2017adversarial}. Adversarial training is also adopted to help learn better feature representations~\cite{liu2017adversarial}. EasyTransfer features Domain-Relationship Specific Shared (DRSS) method to jointly learn the shared feature representations and domain relationships in a unified model.

\boldpara{Instance-based methods.} This type of methods seeks to re-weight the source samples so that data from the source domain and the target domain would share a similar data distribution \cite{chen2011co,huang2007correcting,ruder2017learning,qu2019learning,mgtl}. 
The TL module is typically considered as a sub-module of the data selection framework~\cite{ruder2017learning}. The studies in~\cite{qu2019learning,mgtl} consider building a reinforced selector to help select high-quality source data to help the target.

\boldpara{Model-based methods.} These methods try to transfer the knowledge from the model itself and usually are applied by distilling a big model to a small model~\cite{hinton_kd,DBLP:journals/corr/RomeroBKCGB14,DBLP:conf/ijcai/ChenLQWLDDHLZ20}. Recently, the research of knowledge distillation (KD) from a large pretrain language model becomes popular. \citet{DBLP:journals/corr/abs-1903-12136} distills BERT into BiLSTM networks and achieve comparable results. DistilBERT ~\citep{DBLP:journals/corr/abs-1910-01108} applies KD loss in the pre-training stage, while BERT-PKD~\citep{DBLP:journals/corr/abs-1908-09355} distills BERT into shallow Transformers in the fine-tuning stage. TinyBERT~\citep{DBLP:journals/corr/abs-1909-10351} further applies the distillation of hidden attention matrices and embedding matrices and distills BERT with a two-stage KD process. AdaBERT ~\citep{DBLP:conf/ijcai/ChenLQWLDDHLZ20} uses neural architecture search to adaptively find small architectures for different tasks. EasyTransfer develops the MetaKD method to use meta-teacher learning and meta-distillation to digest transferable knowledge across domains while knowledge distillation and achieves better results on cross-domain tasks.

\boldpara{Meta learning.} This type of methods is slightly different from previous TL categories as meta learning does not directly transfer knowledge from source domains to target ones. Instead, it aims at learning ``meta-learners'' that can digest knowledge from various tasks~\cite{DBLP:journals/corr/abs-2004-11149,DBLP:journals/corr/abs-2004-05439}. After the training of ``meta-learners'', only a few fine-tuning steps are sufficient for ``meta-learners'' to adapt to these tasks with high performance. In EasyTransfer, we treat meta learning as a special type of TL method and design cross-domain and cross-task training algorithms to capture the transferable knowledge.




\section{EasyTransfer Infrastructure}
\label{sec:infra}

\begin{figure}[t!]
    \centering
	\includegraphics[width=1.0\linewidth]{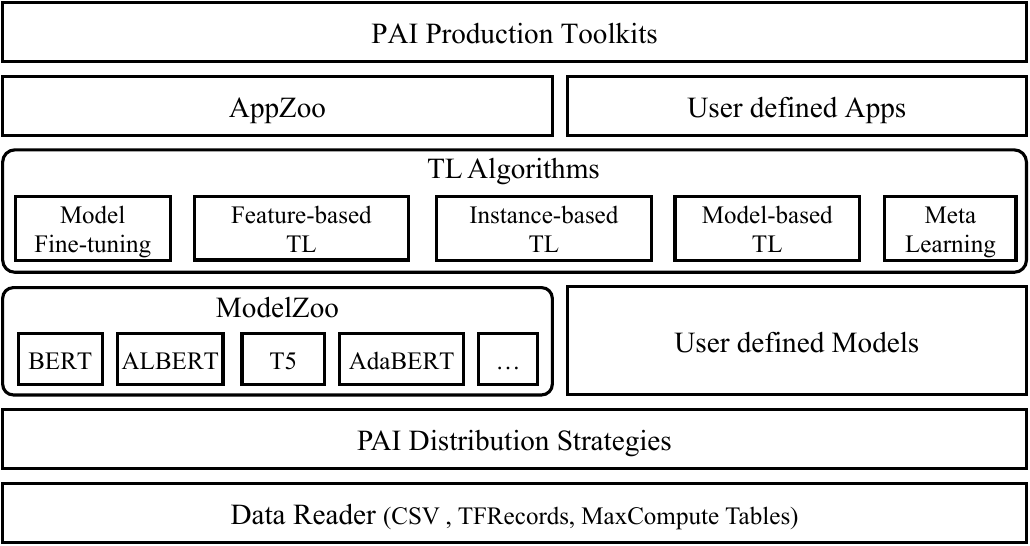}
	\caption{An overview of the EasyTransfer toolkit.}\label{fig:overview}
\end{figure}

In this section, we provide an overview of the EasyTransfer toolkit. The high-level framework of EasyTransfer is shown in Figure~\ref{fig:overview}. EasyTransfer supports rich data readers to process data from multiple data sources and integrates the efficient distribution strategies from Alibaba Platform of AI (PAI). Based on this, users can either the pre-trained models from the ModelZoo or use EasyTransfer APIs to build their own models. EasyTransfer supports the mainstream TL algorithms and a set of in-house developed TL algorithms. It is also equipped with the PAI production toolkit to make it easy for online deployment.

\subsection{Data Storage}
Besides data storage in TFRecord or in the vanilla CSV format, EasyTransfer recommends storing data in MaxCompute~\footnote{https://www.alibabacloud.com/product/maxcompute} tables for high-performance data reading and caching to improve model training efficiency. The MaxCompute platform has server layers to split a data reading job into multiple sub-jobs automatically to support data sharding for distributed training and prediction. Details of resource scheduling in MaxpCompute can be found in~\cite{fuxi}.


\subsection{Distributed Training Strategies}\label{sec:dist}
EasyTransfer leverages Whale, a distributed training framework developed by PAI, that facilitates the distributed training of industrial-level giant models.
Whale provides multiple parallelism strategies including Data Parallelism (DP), Pipeline and a hybrid strategy that combines both DP and Pipeline. 

\boldpara{Whale DP.} The key difference between Whale DP and standard DP lies in the optimization of communication burden. We find that both intra-machine (between multiple local GPUs) and inter-machine (between workers) communications account for a high proportion of running time, especially for large models. As the intra-machine communication bandwidth is much higher than inter-machine bandwidth, inter-machine communication becomes a bottleneck for high-speed training. To alleviate this, we consider hierarchical collective communication to first perform intra-machine data aggregation and then inter-machine aggregation. This helps to reduce the inter-machine communication burden. Furthermore, we consider fusing gradients of different tensors together to reduce communication overhead. We first estimate the generation time of different tensors in the static model graph and then fuse those tensors with similar generation time together for communication. Last but not least, Whale also supports deep gradient compression~\cite{lin2017deep} and half-precision compression.

\boldpara{Whale HP.} When it comes to super large model training where the model cannot be hosted in a single GPU, the above DP strategy is not sufficient. To support this, Whale considers a Hybrid Parallelism (HP) strategy to combine DP with model parallelism and Pipeline. 
The key idea in model parallelism is to partition large model nicely to different devices so that each device has a workload up to its own memory and computation capability. Since a model is split into different devices, there is a dependency between devices when performing feed-forward and backward operations. Pipeline is further used to reduce the waiting time between devices in these scenarios.
Whale considers an autograph partition module to partition a model according to the Floating-Point Operations Per Second (FLOPS) and memory cost of each stage of model training. By cooperating with auto gradient checkpointing, Whale could balance the FLOPs and memory cost among pipeline stages on different devices and improve the overall training efficiencies for large models. By default, we consider low-precision weights (FP16) for communication to achieve better training efficiency.

By using Whale, EasyTransfer serves users with more efficient and powerful distributed training capabilities. 
EasyTransfer is also equipped with our AI Compiler optimizations, it is observed that memory access volume and framework overhead can be saved significantly, details can be found in \cite{zheng2020fusionstitching}.

\subsection{ModelZoo \& AppZoo}

To help users better develop NLP models and applications, EasyTransfer provides a rich family of pretrained models in ModelZoo and a comprehensive NLP applications in AppZoo.

\boldpara{ModelZoo.}
PLMs such as BERT~\cite{devlin2019bert}, RoBERTa~\cite{liu2019roberta}, T5~\cite{raffel2020exploring} and GPT2~\cite{radford2019language} have been the most successful deep learning models for NLP. 
To help users to train models with PLMs, we build ModelZoo in EasyTransfer, which offers pre-trained models including mainstream PLMs including all mainstream transformer-based PLMs, together with the cross-modality model FashionBERT~\cite{fashionbert}.
Our ModelZoo is fully compatible with pre-trained models from open-source toolkits such as Huggingface Transformers.

Table~\ref{fig:modelzoo} presents pre-trained models that EasyTransfer offers to the public with different model sizes ranging from tiny to xxLarge. Note that, Base models have the hidden size of 768, 12 layers, and 12 attention heads, Large models have the hidden size of 1024, 24 layers, and 16 attention heads, xLarge models have the hidden size of 2048, 24 layers, and 32 attention heads, and xxLarge models have the hidden size of 4096, 12 layers and 64 attention heads.

We will also release our pre-trained super-large scale models (with up to 100 billion parameters) in the near future.

\begin{table}
 \centering
\caption{The EasyTransfer ModelZoo. cn means models pre-trained over the Chinese corpus. en means models pre-trained over the English corpus.}
\label{fig:modelzoo}
\begin{tabular}{l c c c c c}
  \toprule 
  ModelZoo & Tiny & Base & Large & xLarge & xxLarge \\ 
  \midrule 
  BERT-cn/en & \checkmark & \checkmark & \checkmark & & \\
  ALBERT-cn/en & \checkmark & \checkmark & \checkmark & \checkmark & \checkmark \\
  RoBERTa-cn/en & \checkmark & \checkmark & \checkmark & & \\
  T5-cn/en & \checkmark & \checkmark & \checkmark & & \\
 \bottomrule 
\end{tabular}
\end{table}

\boldpara{Random Sparse Attention (RSA).} 
The key innovation in transformer-based PLMs is the self-attention mechanism, which is highly effective in capturing the relationship between tokens in an input sequence.
However, 
the attention computation and memory consumption grow quadratically with the sequence length $n$, which limits their application to real-world scenarios that require long sequence modeling such as document modeling and cross-modality learning.
To address this limitation, we develop a novel sparse attention kernel that reduces memory footprints of attention computation via random block-based sparse multi-head attention. More specifically, our approach first blockifies the attention matrices, randomly selects some blocks and then distributes the blocks over multiple attention heads. 
Our random sparse attention decreases the memory footprints and computation FLOPs from $O(n^2)$ to $O(n^2/k)$ (where $k$ is the number of the local attention blocks), and keeps the model quality comparable with the full-attention pre-trained models.

\boldpara{AppZoo.}
To help users better develop NLP applications with our toolkit, we further provide a comprehensive NLP application tool AppZoo.
It supports running applications with a few command-line arguments and provides 10+ mainstream NLP application models for users.
The AppZoo provides rich modules for users to build different application pipelines, supporting four classes of NLP applications, including text classification, text matching, machine reading comprehension, and sequence labeling, with more than 10 models. The detailed application list is at Table~\ref{tab:applications}.

\begin{table}[t!]
    \centering
    \caption{TL algorithms supported by EasyTransfer. Algorithms marked as ``new'' are in-house developed methods.}
    \label{tab:algorithms}
    \begin{tabular}{l l c}
        \toprule
        Type & Algorithms & New \\
        \midrule
        Model Fine-tuning & ModelZoo~(BERT, GPT, etc.) &   \\
        & FashionBERT~\cite{fashionbert} & $\checkmark$ \\\midrule
        Feature-based TL & FullyShared~\cite{mou:EMNLP2016} &  \\
            & SharedPrivate~\cite{liu2017adversarial} &   \\
            & SharedPrivate-Adv~\cite{liu2017adversarial} &   \\
            & DRSS~\cite{drss} &  $\checkmark$ \\
            & DRSS-Adv~\cite{drss} &  $\checkmark$ \\\midrule
        Instance-based TL & Ruder \& Plank~\cite{ruder2017learning} & \\
            & RTL~\cite{qu2019learning} & $\checkmark$ \\
            & MGTL~\cite{mgtl} & $\checkmark$ \\\midrule
        Model-based TL & BERT-PKD~\cite{DBLP:journals/corr/abs-1908-09355} &  \\
            & DistillBERT~\cite{DBLP:journals/corr/abs-1910-01108} &  \\
            & TinyBERT~\cite{DBLP:journals/corr/abs-1909-10351} & \\
            & AdaBERT~\cite{DBLP:conf/ijcai/ChenLQWLDDHLZ20} & $\checkmark$ \\
            & MetaKD~\cite{DBLP:conf/acl/Pan0QZLH20} & $\checkmark$ \\\midrule
        Meta Learning & MetaFT~\cite{emnlp2020mft} & $\checkmark$ \\
            & MetaDTL & $\checkmark$ \\
        \bottomrule
    \end{tabular}
    \label{tab:algorithms}
\end{table}
\begin{table}
    \centering
    \caption{The model list of the EasyTransfer AppZoo. Here, ``\textbf{ModelZoo}'' means that one can use any PLMs from ModelZoo such as BERT, GPT, etc. for fine-tuning on down-streaming tasks.}
    \label{tab:applications}
    \begin{tabular}{l r}
        \toprule
        Application Type & Model \\
        \midrule
        \multirow {2}{*}{Text Classification} &  ModelZoo \\
         & TextCNN \\
        \midrule
        \multirow {3}{*}{Text Matching} &  ModelZoo \\
         & BiCNN, HCNN \\
         & DAM, DAM+ \\
        \midrule
        \multirow {2}{*}{Machine Reading Comprehension} &  ModelZoo \\
         & HAE \\
        \midrule
        \multirow {3}{*}{Sequence Labeling} &  ModelZoo \\
         & BERT-CRF \\
         & BERT-BiLSTM-CRF \\
        \midrule
    \end{tabular}
\end{table}

\subsection{Production Ready Platform}
One advantage of the EasyTransfer platform is that it can directly generate production-ready models for efficient online prediction. After a training procedure is completed, a serialization model in the SavedModel format can be automatically generated from the checkpoint with the highest validation performance. As a default setting, EasyTransfer allows users to deploy the models on PAI Elastic Algorithm Service (EAS), which is an online prediction service that supports deploying machine learning models as RESTful APIs based on heterogeneous hardware (either CPUs or GPUs). By sending simple HTTP requests, users can use the online model service with high concurrency and stability. It is worth mentioning that users are free to use other online prediction services as well.

\section{EasyTransfer Algorithms}\label{sec:algos}

Recall that, the diverse real-world applications require different types of TL algorithms, yet there are no TL toolkits available for users to examine the state-of-the-art TL algorithms and also develop new TL algorithms. To bridge this gap, EasyTransfer not only integrates the main-stream TL algorithms but also provides a comprehensive suite of in-house developed TL algorithms for users to explore. Table~\ref{tab:algorithms} provides an overview of algorithms supported in EasyTransfer  (those algorithms marked as ``new'' in the table are newly developed methods). 
With EasyTransfer, users can either develop their own algorithms using the build-in APIs or directly use the in-house developed algorithms.


\subsection{\`A la Carte Algorithms}
Based on the above infrastructure, EasyTransfer provides both low-level and high-level layer APIs for users to build their own algorithms (\`a la carte).
The layers include basic deep learning layers such as \textit{dense}, \textit{linear} and \textit{LSTM}, NLP layers such as \textit{BERT} and \textit{Transformer}, and Convolution (CV) layers such as \textit{Conv} and \textit{Flatten}. These layers can be also combined with standard layers in Tensorflow\footnote{https://github.com/tensorflow}. Users can use pre-trained models from ModelZoo to build their applications. For instance, ``bert\_base\_en'' is one of the pre-trained models in ModelZoo (i.e., the BERT-base model for English). The detailed list of pre-trained models that we provide can be found in the documentation~\footnote{https://www.yuque.com/easytransfer/cn/geiy58\#IKdVp}.
Furthermore, users can use pre-defined layers or standard Tensorflow layers to build their models. For example, Code~\ref{case1} shows an example of building a BERT-based fine-tuning model for text classification.

\begin{figure}[!t]
\begin{minipage}{0.48\textwidth}
\begin{lstlisting}[language=python,caption=Model fine-tuning example., label=case1]
class TextClassification(base_model):

    def build_logits(self, features, mode=None):
        model = model_zoo.get_pretrained_model('bert_base_en')
        dense = layers.Dense(self.num_labels)
        _, output = model([features], mode=mode)
        return dense(output), label_ids

    def build_loss(self, logits, labels):
        return softmax_cross_entropy(labels, self.num_labels, logits)
    
    def build_eval_metrics(self, logits, labels):
        return classification_eval_metrics(logits, labels, self.num_labels)
\end{lstlisting}
\end{minipage}
\end{figure}


    	

\subsection{In-house Developed Algorithms}

EasyTransfer covers TL algorithms in all the five directions of TL namely model fine-tuning, feature-based TL, instance-based TL, model-based TL, and meta learning.
All these algorithms have been tested in real-world applications and deployed online to support real-world business.

\boldpara{Model Fine-tuning.} 
The most widely used TL algorithm for PLMs is model fine-tuning.
For example, a few fine-tuning steps on BERT and T5 can achieve remarkable results for many NLP applications\cite{devlin2019bert}.
We have also provided a wide range of language models pre-trained using our collected datasets and based on the PAI platform.
As for the cross-modality model, we have developed the FashionBERT model~\cite{fashionbert} for the fashion domain. It has been deployed in Alibaba and has proved highly effective in the tasks of item representation learning and item search.

\boldpara{Feature-based TL.} These methods seek to locate a common feature space that can reduce the differences between source and target domains, by transforming the features from one domain to be closer to another, or projecting different domains into a common latent space where the feature distributions are close \cite{shen2017wasserstein}. Besides these algorithms, EasyTransfer features the Domain-Relationship Specific Shared (DRSS) algorithm to simultaneously learn shared representations and domain relationships in a unified framework~\cite{drss}. DRSS can capture the inter-domain and intra-domain relationships within the specific-shared (or shared-private) architecture. Furthermore, DRSS can be coupled with adversarial learning~\cite{liu2017adversarial} to further boost the model performance.

\boldpara{Instance-based TL.} Due to the domain difference, a vanilla TL method may suffer from the negative transfer. Instance-based TL methods seek to mitigate negative transfer by re-weighting source samples so that data from the source domain and the target domain would share a similar data distribution~\cite{chen2011co,huang2007correcting,ruder2017learning}. 
The TL module is typically considered as a sub-module of the data selection framework~\cite{ruder2017learning}. Therefore, the TL module needs to be retrained repetitively to provide sufficient updates to the data selection framework which may suffer from a long training time when applied to neural TL models. 

EasyTransfer supports two in-house developed instance-based TL algorithms, namely Minimax Game-based TL (MGTL)~\cite{mgtl} and Reinforced TL (RTL)~\cite{qu2019learning}. RTL is the first work to leverage reinforcement learning to select high-quality source data to help the TL process.
The reinforced data selector serves as an agent to interact with the environment created by the TL module. The selector ``acts'' on source data instances to select a subset of source data, and feeds them together with target data to the TL environment. The performance gap serves as a delayed reward to update the RL policy. MGTL addresses the sparse reward issue in RTL by providing an additional domain discriminator to help the TL module learn domain-invariant features and also providing immediate rewards to guild the RL policy. In the experiments, we find both RTL and MGTL are shown to more efficient than the basic instance-based deep TL method such as~\cite{ruder2017learning}.

\boldpara{Model-based TL.} Model-based TL, especially learning a light student model using knowledge distillation, is an important aspect of TL for real-time deployment. EasyTransfer is equipped with many knowledge distillation methods~\cite{hinton_kd,DBLP:journals/corr/abs-1910-01108,DBLP:journals/corr/abs-1908-09355,DBLP:journals/corr/abs-1909-10351,DBLP:conf/ijcai/ChenLQWLDDHLZ20} to compress a big model (e.g. 12-layer BERT) to a small model (e.g. 2-layer BERT or CNN). Furthermore, we develop the task-adaptive BERT compression algorithm named AdaBERT~\cite{DBLP:conf/ijcai/ChenLQWLDDHLZ20} with differentiable neural architecture search, which achieves significant inference speedup and model size reduction.


One disadvantage of previous studies is that they mostly focus on single-domain only, which ignores the transferable knowledge from other domains. 
We develop a meta-knowledge Distillation (MetaKD) framework~\cite{DBLP:conf/acl/Pan0QZLH20} to build a meta-teacher model that captures transferable knowledge across domains and passes such knowledge to students. 
Experiments on public multi-domain NLP tasks show the effectiveness and superiority of the proposed MetaKD framework.

\boldpara{Meta Learning.} Apart from the above deep TL algorithms, EasyTransfer is equipped with the ability of meta learning to improve the performance of domain-level and task-level knowledge transfer for large-scale PLMs. For example, the MetaFT algorithm~\cite{emnlp2020mft} is proposed to learn a ``meta-learner'' based on PLMs, aiming to solve a group of NLP tasks across different domains. 


Additionally, we further consider the situation where there exist several similar NLP tasks related to distant domains with different class label sets. The Meta Distant Transfer Learning (MetaDTL) algorithm is proposed and integrated into EasyTransfer to learn the cross-task ``meta-knowledge'' by modeling the implicit relations among multiple tasks and classes, and then selectively learning the task-agnostic cross-task meta-knowledge. After the ``meta-learner'' is obtained, the model can be quickly adapted to a specific task with better performance by model fine-tuning.
Experiments on public datasets show that MetaFT and MetaDTL are capable of learning cross-domain and cross-task knowledge for downstream tasks.

\begin{table}[t!]
\centering
\caption{The split of multiple domains in MNLI and Amazon Review datasets for evaluating MetaKD and MetaFT.} 
\label{table:mnli}
\begin{tabular}{l c c c c}
\toprule 
 Dataset & Domain  & \#Train & \#Dev & \#Test  \\ 
\midrule 
\multirow{5}{*}{MNLI} & Fiction & 69,613 & 7,735 & 1,973 \\
   & Government & 69,615 & 7,735 & 1,945 \\
  & Slate & 69,575 & 7,731 & 1,955 \\
  & Telephone & 75,013 & 8,335 & 1,966 \\
  & Travel & 69,615 & 7,735 & 1,976 \\
  \midrule
  \multirow{4}{*}{Amazon Review}
  & Book & 1,631 & 170 & 199 \\
  & DVD & 1,621 & 194 & 185 \\
  & Electronics & 1,615 & 172 & 213 \\
  & Kitchen & 1,613 & 184 & 203 \\
 \bottomrule 
\end{tabular}
\end{table}

\section{Experiments}
\label{sec:exp}
In this section, we empirically examine the effectiveness and efficiency of the EasyTransfer toolkit in both open datasets and industrial-scale applications.

\subsection{Tasks and Datasets} \label{app.data}
In the experiments, we have examined the EasyTransfer framework over the following tasks and datasets:
\begin{itemize}
\item \textbf{Paraphrase Identification.} This is a task to examine the relationship, i.e., a paraphrase or not, between two input text sequences.
We treat the Quora question pairs~\footnote{\url{https://www.kaggle.com/c/quora-question-pairs}} as the source domain and a paraphrase dataset made available in CIKM AnalytiCup 2018~\footnote{\url{https://tianchi.aliyun.com/competition/introduction.htm?raceId=231661}} as the target. 
The former is a large-scale dataset that covers a variety of topics, while the latter consists of question pairs in the E-commerce domain. We follow the study in~\cite{qu2019learning} for data pre-processing.

\item \textbf{Sentiment Analysis.} We use a popular sentiment analysis dataset Amazon Reviews~\citep{DBLP:conf/acl/BlitzerDP07}, which is widely used in multi-domain text classification tasks. The reviews are annotated as positive or negative. For each domain, there are 2,000 labeled reviews. We randomly split the data into training, development, and testing sets. 

\item \textbf{Text Matching.} For text matching, we use two open text matching datasets to evaluate our method: MNLI~\cite{MultiNLI} and SciTail~\cite{DBLP:conf/aaai/KhotSC18}. MNLI is a large crowd-sourced corpora for textual entailment recognition with diverse text sources. SciTail is a recently released textual entailment dataset in the science domain. The SciTail dataset is smaller than MNLI but with more diversity in terms of linguistic variations. 
Note that each sample in MNLI is a sentence pair that contains a premise, a hypothesis, and a relation label of ENTAILMENT, NEUTRAL, or CONTRADICTION. However, the labels in SciTail only consist of ENTAILMENT and NEUTRAL.

\item \textbf{Item Recommendation.} We have collected one-week user click-through data from two E-commerce websites, namely Taobao and Qintao.
These two platforms share similar users and item features, but data distributions are different. As for the training data, Taobao has 52000 million samples and Qintao has 350 million samples.
\end{itemize}

The data statistics of MNLI and Amazon Reviews for evaluating MetaKD and MetaFT are shown in Table~\ref{table:mnli}.

\begin{figure}[t!]
\centering
\includegraphics[width=.85\linewidth]{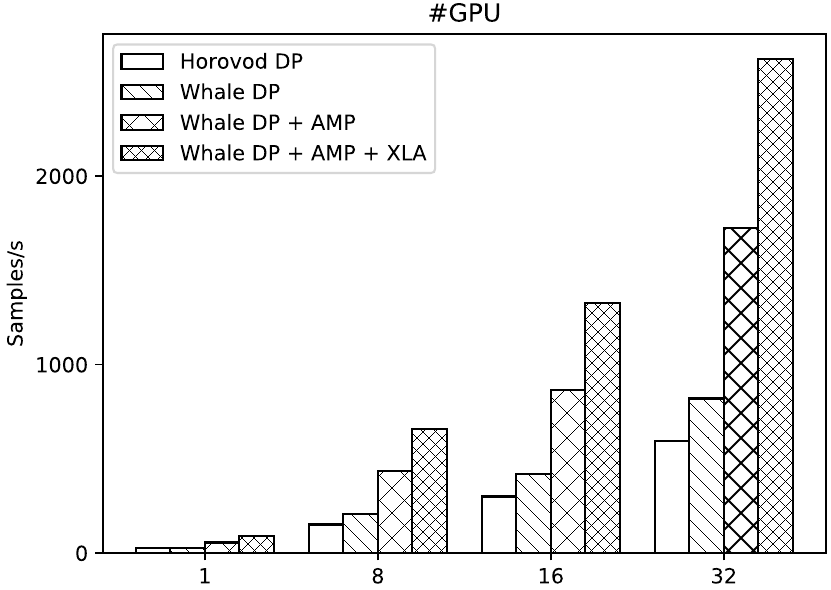}
\vspace{-0.5cm}
\captionof{figure}{Comparison between Horovod DP and Whale DP \\(w. or w/o AMP and XLA).}
\label{fig:albert}
\end{figure}


\subsection{Distributed Training Performance}

\subsubsection{Efficient Data Parallelism}
We use the ALBERT xxLarge model to conduct distributed training using Data Parallelism (DP). We compare our default distributed training engine Whale with an efficient distributed deep learning training framework Horovod~\footnote{https://github.com/horovod/horovod} in Figure ~\ref{fig:albert}. Clearly, the Whale DP has better performance than Horovod DP, which shows the effectiveness of the proposed communication optimization as described in Section~\ref{sec:dist}. Furthermore, EasyTransfer supports auto mixed precision (AMP) and Accelerated Linear Algebra (XLA) which can help to further improve the total throughput (from around 1000 to 2500). As a result, our final DP method achieves a 4.4 times speedup compared to the community version of distributed training.


\begin{figure}[t!]
\centering
\includegraphics[width=.85\linewidth]{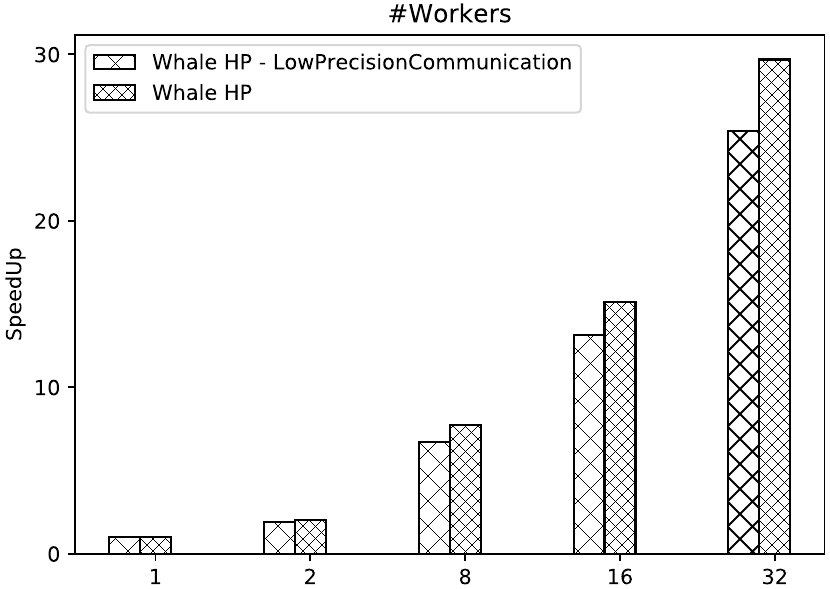}
\vspace{-0.5cm}
\captionof{figure}{EasyTransfer for super large-scale model training (10 billion parameters).}
\label{fig:10B}
\end{figure}



\subsubsection{Efficient Hybrid Parallelism for Super Large Models}
When it comes to super large model pre-training, memory pressure becomes the main issue. For example, the T5-11B model, which consists of 24 layer encoder and decoder, intermediate size 65536 with 128-headed attention producing a model with about 11 billion parameters that can't be fit into one single GPU. Therefore, we use Whale HP to split the model into 8 GPUs within a single worker and train it with multiple workers. As shown in Figure~\ref{fig:10B}, if we don't use low-precision weights (FP16) for communication, we observe a good speedup w.r.t. the number of workers. With low-precision communication, our method achieves a near-liner speedup, e.g. around 30 times speedup with 32 workers.
This shows our framework is with a highly efficient training performance for super-large models.
Last, Whale also supports Mixture-of-Experts based model sharding~\cite{shazeer2017outrageously}, which can train models with up to 100 billion parameters.
\begin{table}[t!]
 \centering
 \small
 \caption{Performance of PLMs on the SuperGLUE benchmark. Bold texts denote best results in each column. ``g-'' means the pre-trained weights are from ``google-bert''.}
 \begin{small}
 \begin{tabular}{l c c c c c c }
 \toprule
 \multirow{2}{*}{Model} & \multicolumn{5}{c}{SuperGLUE Datasets} &  \multirow{2}{*}{Avg}  \\\cline{2-6}
 & CB & COPA & BoolQ & WiC & WSC \\
  \midrule 
  g-bert-base & 75.0 & 61.0 & 74.5 & 69.1 & 63.5 & 68.6 \\
  g-bert-large & 73.2 & 62.0 & 79.1 & 69.8 & \textbf{65.4} & 69.9\\  
  g-albert-large & 85.7 & 68.0 & 79.2 & 72.7 & 63.5 & 73.8 \\ 
  g-albert-xxlarge & 83.9 & \textbf{85.0} & 84.5 & \textbf{75.2} & 63.5 & 78.4\\  \midrule
  pai-albert-xxlarge & \textbf{85.7} & 84.0 & \textbf{85.4} & 74.6 & 63.5 & \textbf{78.6}\\
 \bottomrule
 \end{tabular}
 \end{small}
 \label{tab:glue_results}
 \end{table}

\subsection{PLM Performance}
\boldpara{Benchmark.} In order to evaluate the reliability of our PLMs in EasyTranfer, we evaluate five downstream tasks from the SuperGLUE benchmark. For all tasks, we use 
a limited hyper-parameter search, with batch sizes in \{$8$, $16$, $32$\} and learning rate in \{$1e-5$, $3e-5$, $5e-5$\}. 
We perform early stopping based on each task's evaluation metric on the development set. In this setting, we report the median development set results for each task over three random initialization.  
Table~\ref{tab:glue_results} shows the performance of baseline models on SuperGLUE benchmarks. As we can see, pai-albert-xxlarge can get comparable or better performance than others. This shows the EasyTransfer framework is reliable and can pre-train models with similar performance compared to the open-source framework.

\boldpara{RSA-based Transformers.} We further examine the effectiveness of our proposed RSA-based method on GLUE tasks in Table~\ref{tab:glue_results2}. We find that we can save approximately $25\%$ of the original training time, from 143 hours to 105 hours for pre-training BERT and from 180 to 135 for pre-training T5. It is due to the fact that RSA with a sparse ratio $1/4$ (k=4) saves memory dramatically. Hence it takes less training time to converge to a reasonable loss. 
Both RSA-BERT and RSA-T5 have a similar downstream performance to that of the original BERT and T5 models, respectively. It shows that the RSA approach is generic to the transformer-based architecture. In a nutshell, the RSA method reduces the memory footprint and computation FLOPs and also keeps the model quality comparable with the original transformer-based models.
 
\begin{table}[t!]
\centering
\caption{Performance of PLMs on the CLUE benchmark. Here, ``hour'' means the time for model pre-training.}
 \vspace{-.5em}
\label{tab:glue_results2}
\begin{tabular}{l c c c c c l}
\toprule
\multirow{2}{*}{Model}  & \multicolumn{4}{c}{GLUE Datasets}
& \multirow{2}{*}{Avg}  & \multirow{2}{*}{Hour}\\\cline{2-5}
	&	SST-2	&	QQP	&	MRPC	&	RTE	&	&		\\
\midrule 
BERT 	&	92.1 	&	90.6 	&	85.8 	&	67.2 	&	83.9 	&	143	\\
RSA-BERT$_{k2}$	&	92.2 	&	90.3 	&	86.5 	&	65.7 	&	83.7 	&	137~(-4\%)	\\
RSA-BERT$_{k4}$	&	91.7 	&	90.1 	&	84.8 	&	66.1 	&	83.2 	&	\textbf{105}~(-27\%)	\\\midrule
T5 	&	92.6 	&	90.6 	&	86.0 	&	67.5 	&	84.2 	&	180	\\
RSA-T5$_{k2}$	&	92.1 	&	90.5 	&	86.1 	&	66.7 	&	83.9 	&	170~(-6\%)	\\
RSA-T5$_{k4}$	&	90.3 	&	89.4 	&	86.0 	&	66.0 	&	82.9 	&	\textbf{135}~(-25\%)	\\
\bottomrule
\end{tabular}
\end{table}

\boldpara{Online Applications.} The RSA-based transformers have been widely used in real-world applications inside Alibaba. Most notably, RSA-BERT has been applied on query-item retrieval in Alibaba.com. The key task is to match queries and item titles in the e-commerce search engine of Alibaba.com. As a result, RSA-BERT outperforms the online production method (a DNN based query-item matching method) and has improved around 3\% in terms of AUC. 

\subsection{Evaluations of Feature-based TL}
We evaluate the performance of feature-based TL algorithms on Paraphrase Identification (PI) ~\citep{drss} in Table~\ref{tab:quoraresult}. We can observe that for all the five target domains, Src-Only performs much worse than Tgt-Only. The average performance of Mixed is even worse than Tgt-Only.
This implies that the source domain is quite different from all the target domains. Simply mixing the training data in two domains may lead to the model overfitting to the large domain.
In addition, it is observed that the widely used model fine-tuning method performs slightly better than Tgt-Only in most cases, which shows that pre-training the model on the source domain is beneficial.
Moreover, in all the five domains, the performance of two existing transfer learning frameworks FS and SS are both 1.9\% better than that of Tgt-Only, which proves their usefulness.
Furthermore, our featured methods DRSS and DRSS-Adv can achieve better results than other methods, which shows the importance of domain-relationship learning in feature-based TL. 

\boldpara{Online Applications.} We deploy DRSS-Adv online in the AliMe assistant chatbot~\cite{Li2017AliMe}. The original question-answering module in AliMe is a GBDT model with a set of semantic matching features including TF-IDF scores, word2vec scores, TextCNN scores, etc. We propose to train GBDT by treating the prediction score of our DRSS model as another feature, with the resulting method as GBDT+DRSS. For online evaluation, we randomly sample 2750 questions, where 1317 questions are answered by GBDT and 1433 questions are answered by GBDT-DRSS. We asked business analysts to annotate if the nearest question returned by models expresses the same meaning as the query question. The Precision@1 of GBDT-DRSS is 18.8\% higher than that of GBDT (53.9\% vs. 68.1\%). 


\begin{table}[!tp] 
\centering
\caption{Results of feature-based TL methods on Paraphrase Identification task.}
\vspace{-.5em}
\label{tab:quoraresult}
\begin{tabular}{llllll}
\toprule
Method           & Prec@1 & Rec@1 & F1@1  & ACC     & AUC     \\ \midrule
Tgt-Only    & 71.7  & 55.1 & 62.3 & 79.2   & 82.2   \\
Src-Only    & 61.9  & 36.8 & 46.1 & 71.9   & 68.6   \\
Fine-Tune   & 71.3  & 56.7 & 63.2 & 79.0   & 82.5   \\
FS~\cite{mou:EMNLP2016} & 73.4  & 59.5 & 65.7 & 79.7   & 83.1 \\
SS~\cite{liu2017adversarial} & 74.4  & 60.1 & 66.5 &80.0 & 83.7 \\
SS-Adv~\cite{liu2017adversarial} & 74.3  & 60.3 & 66.6 & 80.8   & 84.2   \\\midrule
DRSS        & \textbf{75.7}  & 60.8 & 67.4 & \textbf{81.2}   & 84.7  \\
DRSS-Adv    & 75.3  & \textbf{62.0} & \textbf{68.0} & 80.9  & \textbf{84.9}   \\\bottomrule
\end{tabular}
\end{table}

\begin{table}[t!]
\caption{Evaluation of instance-based TL methods.}
\label{tab:item_results}
\vspace{-.5em}
\begin{tabular}{l c c c c }
\toprule
\multirow{2}{*}{Method} & \multicolumn{2}{c}{Item Recommendation} & \multicolumn{2}{c}{Text Matching} \\ \cmidrule{2-5}
 & ACC        & AUC  & ACC        & AUC \\ \midrule
Src-only     & 89.1     & 69.1 & 71.1 & 70.9     \\
Tgt-only     & 90.1     & 70.3 & 73.0 & 76.6     \\
TL Method~\cite{mou:EMNLP2016,mou:acl2016}    & 91.0     & 72.1 & 74.5 & 80.4 \\ 
Ruder and Plank~\cite{ruder2017learning}  & 91.2 & 73.3 & 75.2 & 80.6 \\  
RTL  & 91.3 & 73.3 & 76.7 & 81.6 \\  
MGTL & \textbf{91.5}     & \textbf{74.5} & \textbf{77.8} & \textbf{82.5} \\ \bottomrule
\end{tabular}
\end{table}

\subsection{Evaluations of Instance-based TL}
We evaluate the two in-house intance-based TL algorithms in EasyTransfer, namely MGTL~\cite{mgtl} and RTL~\cite{qu2019learning}.
We also compare these methods with a source-data only method, a target-data only method, representative TL algorithms ~\cite{mou:acl2016,mou:EMNLP2016} and an instance selection method with Bayesian optimization~\cite{ruder2017learning} on the task of item recommendation and text matching as studied in~\cite{qu2019learning} in terms of ACC and AUC. 
As shown in Table~\ref{tab:item_results}, a few important observations are as follows. First, Src-only performs worse than Tgt-only, which means source data is close to but different from target data, i.e. there is a domain shift between the source and target domain. In this case, simply using data from the source domain is not satisfactory. Second, the TL method improves the Tgt-only model which means the TL method can help to leverage information from the source domain to help the target domain. Furthermore, the instance selection method in~\cite{ruder2017learning} can further improve the TL model.
MGTL achieves better performance than the method in~\cite{ruder2017learning} and also the vanilla RTL method, which further shows our immediate rewards and delayed rewards are helpful for the tasks. In all, the comparison with other competing methods shows the advantage of the reinforced selector based TL algorithms.

\boldpara{Online Applications.} We have deployed MGTL on the platform Qintao\footnote{https://qintao.taobao.com} to leverage information from Taobao\footnote{https://taobao.com} for the task of item search, as the platform Taobao has 100x more data instances than Qintao. We compare MGTL with a baseline model which is a degenerated version of our model that does not consider source instance selection. We have run 7-day A/B testing on the click-through-rate of the models. Our method outperforms the online model with 3\%+ improvement on average. This brings to an increase of 1.5\%+ of the total gross merchandise volume.

\subsection{Evaluations of Model-based TL}


We further proceed to evaluate MetaKD~\cite{DBLP:conf/acl/Pan0QZLH20} on Amazon Reviews ~\citep{DBLP:conf/acl/BlitzerDP07}. Table~\ref{table:results-amazon} shows the general testing performance of baselines and MetaKD, in terms of accuracy. Here, BERT-s refers to a single BERT teacher trained on each domain, BERT-mix is one BERT teacher trained on the mix of all domain data and BERT-mtl is one teacher trained by multi-task learning over all domains. For distillation baselines, ``$\rightarrow$ TinyBERT'' means using the KD method described in the TinyBERT paper ~\citep{DBLP:journals/corr/abs-1909-10351} to distill the corresponding teacher model. For each method, the teacher model is a BERT-base (Total Parameters=110M) model and the student model is a BERT-Tiny (Total Parameters=14.5M) model.
Compared to all the baseline teacher models, MetaKD achieves the highest accuracy. Our method significantly reduces the model size while preserving a similar performance. Specifically, we reduce the model size to 7.5x smaller (BERT to TinyBERT) with only a minor performance drop (from 89.9\% to 89.4\%). 

\begin{table}[t!] 
 \centering
 \caption{Evaluation of MetaKD over Amazon reviews (with four domains) in terms of accuracy (\%).} 
 \begin{tabular}{lccccc} 
  \toprule 
  Method  & Books & DVD & Elec. & Kit. & Avg.  \\ 
  \midrule 
$\text{BERT}\text{-s}$ & 87.9 & 83.8 & 89.2 & 90.6 & 87.9  \\
$\text{BERT}\text{-mix}$ & 89.9 & 85.9 & 90.1 & \bf 92.1 & 89.5 \\
$\text{BERT}\text{-mtl}$ & 90.5 & 86.5 & 91.1 & 91.1 & 89.8 \\
Meta-teacher & 92.5 & 87.0 & 91.1 & 89.2 & 89.9 \\
 \midrule
$\text{BERT}\text{-s}$ $\rightarrow$ TinyBERT & 83.4 & 83.2 & 89.2 & 91.1 & 86.7 \\
$\text{BERT}\text{-mix}$ $\rightarrow$ TinyBERT & 88.4 & 81.6 & 89.7 & 89.7 & 87.3 \\
$\text{BERT}\text{-mtl}$ $\rightarrow$ TinyBERT & 90.5 & 81.6 & 88.7 & 90.1 & 87.7 \\
MetaKD & \textbf{91.5} & \textbf{86.5} & \textbf{90.1} & 89.7 & \textbf{89.4} \\
 \bottomrule 
 \end{tabular}
\label{table:results-amazon}
\end{table}


\boldpara{Online Applications.} A typical way of building an efficient and effective model for real-world application is to first fine-tune a large model to achieve good performance first, and then distill it to a smaller efficient model. Inside Alibaba, EasyTransfer has provided a widely-used model distillation service to many applications that require low latency, such as user intent detection in AliMe assistant chatbot, malicious reviews detection on e-commerce platforms, etc.

\subsection{Evaluations of Meta Learning}
The MetaFT algorithm leverages the power of the meta-learner, hence is highly effective in the few-shot learning setting. Take the multi-domain MNLI dataset~\citep{MultiNLI} for an example. For each of the five domains, we only use 5\%, 10\%, and 20\% of the original dataset for model fine-tuning. The prediction accuracy scores with and without the MetaFT algorithm are compared, with BERT-base as the underlying PLM. The results are in Table~\ref{tab:few1}. 

As MetaFT only considers learning meta-learners across similar domains, we consider a more challenging case to evaluate MetaDTL.
In this experiment, we randomly sample only 1\%, 2\%, 5\%, 10\%, and 20\% of the original MNLI training data to fine-tune the BERT model. The SciTail~\cite{DBLP:conf/aaai/KhotSC18} training set is used for knowledge transfer. We list the results on the MNLI testing set with and without MetaDTL training in Table~\ref{tab:few}. The results produced by Meta-FT and single-task fine-tuning are also compared. As seen, MetaDTL improves the performance regardless of the percentages of the training sets. It has a larger increase in accuracy on smaller training sets (4.0\% increase on 1\% of the training set vs. 1.0\% increase on 20\%).
In summary, MeteFT and MetaDTL are capable of training meta-learners across domains and tasks, suitable for tasks with very little training data.

\begin{table}[t!]
\centering
\caption{Evaluation of MetaFT on  natural language inference over few-shot MNLI in terms of accuracy (\%).} \label{tab:few1} 
\begin{tabular}{l ll ll ll}  
\toprule
\multirow{2}{*}{Domain} & \multicolumn{2}{c}{MetaFT?} & \multicolumn{2}{c}{MetaFT?} & \multicolumn{2}{c}{MetaFT?}\\
\cline{2-7}
& No & Yes & No & Yes & No & Yes\\
\hline
Ratio of training set & \multicolumn{2}{c}{5\%} & \multicolumn{2}{c}{10\%} & \multicolumn{2}{c}{20\%}\\
\hline
Telephone & 70.5 & 74.7 & 74.1 & 76.4 & 75.9 & 79.8\\
Government & 76.5 & 78.1 & 78.8 & 81.0 & 80.5 & 82.9\\
Slate & 64.2 & 69.8 & 67.6 & 71.8 & 71.8 & 74.1\\
Travel & 71.9 & 75.4 & 74.8 & 78.1 & 78.3 & 80.3\\
Fiction & 69.7 & 73.8 & 73.3 & 76.6 & 76.2 & 78.4\\
\hline
Average & 70.5 & \bf 74.4 & 73.7 & \bf 76.8 & 76.5 & \bf 79.1\\
\bottomrule
\end{tabular}
\end{table}  

\begin{table}[t!]
\centering
\caption{Comparison between MetaFT and MetaDTL on the testing set when only part of the MNLI training set is employed in terms of accuracy (\%).}
\label{tab:few}
\begin{tabular}{l l l l}
\toprule
Percentage & Single-task & MetaFT & MetaDTL\\
\hline
1\% & 62.5 & 64.1 & \bf 66.5 (+4.0\%)\\
2\% & 67.5 & 68.2 & \bf 69.8 (+2.3\%)\\
5\% & 72.8 & 73.8 & \bf 74.2 (+1.4\%)\\
10\% & 75.8 & 76.2 & \bf 77.6 (+1.8\%)\\
20\% & 80.4 & 80.8 & \bf 81.4 (+1.0\%)\\
\bottomrule
\end{tabular}
\end{table}

\boldpara{Online Applications.}
The ``meta-learners'' trained by EasyTransfer are frequently applied in various applications in Alibaba, especially for emerging domains with few training data. For example, in the AliMe assistant chatbot, there are over one thousand similar user intent classification tasks for different domains and businesses. By applying the ``meta-learner'' fine-tuned on PLMs, the overall accuracy is improved by over 8\%, compared to the online system.

\section{Conclusion and Future Work}
\label{conclusion_label}
In this paper, we introduced EasyTransfer, a toolkit that is designed to make it easy to develop deep TL algorithms for NLP applications. To cope with the development of large pretrained models, EasyTransfer is built with a scalable architecture to support large model training and inference. And to meet the need of diverse real-world applications, EasyTrasfer supports a rich family of TL algorithms, and has been used to support many (20+) business scenarios in Alibaba. 
It has been integrated into Alibaba Cloud to support many external business needs. The toolkit has been open-sourced to promote research for deep TL and NLP applications.

\bibliographystyle{ACM-Reference-Format}
\bibliography{sample-base}


\begin{thebibliography}{54}


\ifx \showCODEN    \undefined \def \showCODEN     #1{\unskip}     \fi
\ifx \showDOI      \undefined \def \showDOI       #1{#1}\fi
\ifx \showISBNx    \undefined \def \showISBNx     #1{\unskip}     \fi
\ifx \showISBNxiii \undefined \def \showISBNxiii  #1{\unskip}     \fi
\ifx \showISSN     \undefined \def \showISSN      #1{\unskip}     \fi
\ifx \showLCCN     \undefined \def \showLCCN      #1{\unskip}     \fi
\ifx \shownote     \undefined \def \shownote      #1{#1}          \fi
\ifx \showarticletitle \undefined \def \showarticletitle #1{#1}   \fi
\ifx \showURL      \undefined \def \showURL       {\relax}        \fi
\providecommand\bibfield[2]{#2}
\providecommand\bibinfo[2]{#2}
\providecommand\natexlab[1]{#1}
\providecommand\showeprint[2][]{arXiv:#2}

\bibitem[\protect\citeauthoryear{Beltagy, Peters, and Cohan}{Beltagy
  et~al\mbox{.}}{2020}]%
        {beltagy2020longformer}
\bibfield{author}{\bibinfo{person}{Iz Beltagy}, \bibinfo{person}{Matthew~E.
  Peters}, {and} \bibinfo{person}{Arman Cohan}.}
  \bibinfo{year}{2020}\natexlab{}.
\newblock \showarticletitle{Longformer: The Long-Document Transformer}.
\newblock \bibinfo{journal}{\emph{arXiv:2004.05150}} (\bibinfo{year}{2020}).
\newblock


\bibitem[\protect\citeauthoryear{Blitzer, Dredze, and Pereira}{Blitzer
  et~al\mbox{.}}{2007}]%
        {DBLP:conf/acl/BlitzerDP07}
\bibfield{author}{\bibinfo{person}{John Blitzer}, \bibinfo{person}{Mark
  Dredze}, {and} \bibinfo{person}{Fernando Pereira}.}
  \bibinfo{year}{2007}\natexlab{}.
\newblock \showarticletitle{Biographies, Bollywood, Boom-boxes and Blenders:
  Domain Adaptation for Sentiment Classification}. In
  \bibinfo{booktitle}{\emph{ACL}}.
\newblock


\bibitem[\protect\citeauthoryear{Chen, Li, Qiu, Wang, Li, Ding, Deng, Huang,
  Lin, and Zhou}{Chen et~al\mbox{.}}{2020}]%
        {DBLP:conf/ijcai/ChenLQWLDDHLZ20}
\bibfield{author}{\bibinfo{person}{D. Chen}, \bibinfo{person}{Y. Li},
  \bibinfo{person}{M. Qiu}, \bibinfo{person}{Z. Wang}, \bibinfo{person}{B. Li},
  \bibinfo{person}{B. Ding}, \bibinfo{person}{H. Deng}, \bibinfo{person}{J.
  Huang}, \bibinfo{person}{W. Lin}, {and} \bibinfo{person}{J. Zhou}.}
  \bibinfo{year}{2020}\natexlab{}.
\newblock \showarticletitle{AdaBERT: Task-Adaptive {BERT} Compression with
  Differentiable Neural Architecture Search}. In
  \bibinfo{booktitle}{\emph{IJCAI}}. \bibinfo{pages}{2463--2469}.
\newblock


\bibitem[\protect\citeauthoryear{Chen, Weinberger, and Blitzer}{Chen
  et~al\mbox{.}}{2011}]%
        {chen2011co}
\bibfield{author}{\bibinfo{person}{Minmin Chen}, \bibinfo{person}{Kilian~Q
  Weinberger}, {and} \bibinfo{person}{John Blitzer}.}
  \bibinfo{year}{2011}\natexlab{}.
\newblock \showarticletitle{Co-training for domain adaptation}. In
  \bibinfo{booktitle}{\emph{NeurIPS}}. \bibinfo{pages}{2456--2464}.
\newblock


\bibitem[\protect\citeauthoryear{Child, Gray, Radford, and Sutskever}{Child
  et~al\mbox{.}}{2019}]%
        {child2019generating}
\bibfield{author}{\bibinfo{person}{Rewon Child}, \bibinfo{person}{Scott Gray},
  \bibinfo{person}{Alec Radford}, {and} \bibinfo{person}{Ilya Sutskever}.}
  \bibinfo{year}{2019}\natexlab{}.
\newblock \showarticletitle{Generating Long Sequences with Sparse
  Transformers}.
\newblock \bibinfo{journal}{\emph{arXiv:1904.10509}} (\bibinfo{year}{2019}).
\newblock


\bibitem[\protect\citeauthoryear{Devlin, Chang, Lee, and Toutanova}{Devlin
  et~al\mbox{.}}{2019}]%
        {devlin2019bert}
\bibfield{author}{\bibinfo{person}{Jacob Devlin}, \bibinfo{person}{Ming{-}Wei
  Chang}, \bibinfo{person}{Kenton Lee}, {and} \bibinfo{person}{Kristina
  Toutanova}.} \bibinfo{year}{2019}\natexlab{}.
\newblock \showarticletitle{{BERT:} Pre-training of Deep Bidirectional
  Transformers for Language Understanding}. In
  \bibinfo{booktitle}{\emph{NAACL-HLT}}.
\newblock


\bibitem[\protect\citeauthoryear{Gao, Jin, Chen, Qiu, Li, Wei, Hu, and
  Wang}{Gao et~al\mbox{.}}{2020}]%
        {fashionbert}
\bibfield{author}{\bibinfo{person}{Dehong Gao}, \bibinfo{person}{Linbo Jin},
  \bibinfo{person}{Ben Chen}, \bibinfo{person}{Minghui Qiu},
  \bibinfo{person}{Peng Li}, \bibinfo{person}{Yi Wei}, \bibinfo{person}{Yi Hu},
  {and} \bibinfo{person}{Hao Wang}.} \bibinfo{year}{2020}\natexlab{}.
\newblock \showarticletitle{FashionBERT: Text and Image Matching with Adaptive
  Loss for Cross-modal Retrieval}. In \bibinfo{booktitle}{\emph{SIGIR}}.
\newblock


\bibitem[\protect\citeauthoryear{He, Zhang, Ren, and Sun}{He
  et~al\mbox{.}}{2016}]%
        {he2016deep}
\bibfield{author}{\bibinfo{person}{Kaiming He}, \bibinfo{person}{Xiangyu
  Zhang}, \bibinfo{person}{Shaoqing Ren}, {and} \bibinfo{person}{Jian Sun}.}
  \bibinfo{year}{2016}\natexlab{}.
\newblock \showarticletitle{Deep Residual Learning for Image Recognition}. In
  \bibinfo{booktitle}{\emph{CVPR}}. \bibinfo{pages}{770--778}.
\newblock


\bibitem[\protect\citeauthoryear{Hinton, Vinyals, and Dean}{Hinton
  et~al\mbox{.}}{2015}]%
        {hinton_kd}
\bibfield{author}{\bibinfo{person}{Geoffrey Hinton}, \bibinfo{person}{Oriol
  Vinyals}, {and} \bibinfo{person}{Jeffrey Dean}.}
  \bibinfo{year}{2015}\natexlab{}.
\newblock \showarticletitle{Distilling the Knowledge in a Neural Network}. In
  \bibinfo{booktitle}{\emph{NIPS Deep Learning and Representation Learning
  Workshop}}.
\newblock


\bibitem[\protect\citeauthoryear{Hospedales, Antoniou, Micaelli, and
  Storkey}{Hospedales et~al\mbox{.}}{2020}]%
        {DBLP:journals/corr/abs-2004-05439}
\bibfield{author}{\bibinfo{person}{Timothy~M. Hospedales},
  \bibinfo{person}{Antreas Antoniou}, \bibinfo{person}{Paul Micaelli}, {and}
  \bibinfo{person}{Amos~J. Storkey}.} \bibinfo{year}{2020}\natexlab{}.
\newblock \showarticletitle{Meta-Learning in Neural Networks: {A} Survey}.
\newblock \bibinfo{journal}{\emph{arXiv:2004.05439}} (\bibinfo{year}{2020}).
\newblock


\bibitem[\protect\citeauthoryear{Howard and Ruder}{Howard and Ruder}{2018}]%
        {howard2018universal}
\bibfield{author}{\bibinfo{person}{Jeremy Howard} {and}
  \bibinfo{person}{Sebastian Ruder}.} \bibinfo{year}{2018}\natexlab{}.
\newblock \showarticletitle{Universal Language Model Fine-tuning for Text
  Classification}. In \bibinfo{booktitle}{\emph{ACL}}.
  \bibinfo{pages}{328--339}.
\newblock


\bibitem[\protect\citeauthoryear{Huang, Gretton, Borgwardt, Sch{\"o}lkopf, and
  Smola}{Huang et~al\mbox{.}}{2007}]%
        {huang2007correcting}
\bibfield{author}{\bibinfo{person}{Jiayuan Huang}, \bibinfo{person}{Arthur
  Gretton}, \bibinfo{person}{Karsten Borgwardt}, \bibinfo{person}{Bernhard
  Sch{\"o}lkopf}, {and} \bibinfo{person}{Alex~J Smola}.}
  \bibinfo{year}{2007}\natexlab{}.
\newblock \showarticletitle{Correcting sample selection bias by unlabeled
  data}. In \bibinfo{booktitle}{\emph{NeurIPS}}. \bibinfo{pages}{601--608}.
\newblock


\bibitem[\protect\citeauthoryear{Jiao, Yin, Shang, Jiang, Chen, Li, Wang, and
  Liu}{Jiao et~al\mbox{.}}{2020}]%
        {DBLP:journals/corr/abs-1909-10351}
\bibfield{author}{\bibinfo{person}{Xiaoqi Jiao}, \bibinfo{person}{Yichun Yin},
  \bibinfo{person}{Lifeng Shang}, \bibinfo{person}{Xin Jiang},
  \bibinfo{person}{Xiao Chen}, \bibinfo{person}{Linlin Li},
  \bibinfo{person}{Fang Wang}, {and} \bibinfo{person}{Qun Liu}.}
  \bibinfo{year}{2020}\natexlab{}.
\newblock \showarticletitle{TinyBERT: Distilling {BERT} for Natural Language
  Understanding}. In \bibinfo{booktitle}{\emph{EMNLP (Findings)}}.
  \bibinfo{pages}{4163--4174}.
\newblock


\bibitem[\protect\citeauthoryear{Khot, Sabharwal, and Clark}{Khot
  et~al\mbox{.}}{2018}]%
        {DBLP:conf/aaai/KhotSC18}
\bibfield{author}{\bibinfo{person}{Tushar Khot}, \bibinfo{person}{Ashish
  Sabharwal}, {and} \bibinfo{person}{Peter Clark}.}
  \bibinfo{year}{2018}\natexlab{}.
\newblock \showarticletitle{SciTaiL: {A} Textual Entailment Dataset from
  Science Question Answering}. In \bibinfo{booktitle}{\emph{AAAI}}.
  \bibinfo{pages}{5189--5197}.
\newblock


\bibitem[\protect\citeauthoryear{Lan, Chen, Goodman, Gimpel, Sharma, and
  Soricut}{Lan et~al\mbox{.}}{2020}]%
        {lan2020albert}
\bibfield{author}{\bibinfo{person}{Zhenzhong Lan}, \bibinfo{person}{Mingda
  Chen}, \bibinfo{person}{Sebastian Goodman}, \bibinfo{person}{Kevin Gimpel},
  \bibinfo{person}{Piyush Sharma}, {and} \bibinfo{person}{Radu Soricut}.}
  \bibinfo{year}{2020}\natexlab{}.
\newblock \showarticletitle{{ALBERT:} {A} Lite {BERT} for Self-supervised
  Learning of Language Representations}. In \bibinfo{booktitle}{\emph{ICLR}}.
\newblock


\bibitem[\protect\citeauthoryear{Lee, Lee, Kim, Kosiorek, Choi, and Teh}{Lee
  et~al\mbox{.}}{2019}]%
        {lee2019set}
\bibfield{author}{\bibinfo{person}{Juho Lee}, \bibinfo{person}{Yoonho Lee},
  \bibinfo{person}{Jungtaek Kim}, \bibinfo{person}{Adam~R. Kosiorek},
  \bibinfo{person}{Seungjin Choi}, {and} \bibinfo{person}{Yee~Whye Teh}.}
  \bibinfo{year}{2019}\natexlab{}.
\newblock \showarticletitle{Set Transformer: A Framework for Attention-based
  Permutation-Invariant Neural Networks}.
\newblock \bibinfo{journal}{\emph{arXiv:1810.00825}} (\bibinfo{year}{2019}).
\newblock


\bibitem[\protect\citeauthoryear{Lewis, Liu, Goyal, Ghazvininejad, Mohamed,
  Levy, Stoyanov, and Zettlemoyer}{Lewis et~al\mbox{.}}{2019}]%
        {lewis2019bart}
\bibfield{author}{\bibinfo{person}{Mike Lewis}, \bibinfo{person}{Yinhan Liu},
  \bibinfo{person}{Naman Goyal}, \bibinfo{person}{Marjan Ghazvininejad},
  \bibinfo{person}{Abdelrahman Mohamed}, \bibinfo{person}{Omer Levy},
  \bibinfo{person}{Ves Stoyanov}, {and} \bibinfo{person}{Luke Zettlemoyer}.}
  \bibinfo{year}{2019}\natexlab{}.
\newblock \showarticletitle{BART: Denoising Sequence-to-Sequence Pre-training
  for Natural Language Generation, Translation, and Comprehension}.
\newblock \bibinfo{journal}{\emph{arXiv:1910.13461}} (\bibinfo{year}{2019}).
\newblock


\bibitem[\protect\citeauthoryear{Li, Qiu, Chen, Wang, Gao, Huang, Ren, Zhao,
  Zhao, Wang, Jin, and Chu}{Li et~al\mbox{.}}{2017}]%
        {Li2017AliMe}
\bibfield{author}{\bibinfo{person}{F. Li}, \bibinfo{person}{M. Qiu},
  \bibinfo{person}{H. Chen}, \bibinfo{person}{X. Wang}, \bibinfo{person}{X.
  Gao}, \bibinfo{person}{J. Huang}, \bibinfo{person}{J. Ren},
  \bibinfo{person}{Z. Zhao}, \bibinfo{person}{W. Zhao}, \bibinfo{person}{L.
  Wang}, \bibinfo{person}{G. Jin}, {and} \bibinfo{person}{W. Chu}.}
  \bibinfo{year}{2017}\natexlab{}.
\newblock \showarticletitle{{AliMe Assist : An Intelligent Assistant for
  Creating an Innovative E-commerce Experience}}. In
  \bibinfo{booktitle}{\emph{CIKM}}.
\newblock


\bibitem[\protect\citeauthoryear{Li, Duan, Fang, Gong, Jiang, and Zhou}{Li
  et~al\mbox{.}}{2019}]%
        {li2019unicodervl}
\bibfield{author}{\bibinfo{person}{Gen Li}, \bibinfo{person}{Nan Duan},
  \bibinfo{person}{Yuejian Fang}, \bibinfo{person}{Ming Gong},
  \bibinfo{person}{Daxin Jiang}, {and} \bibinfo{person}{Ming Zhou}.}
  \bibinfo{year}{2019}\natexlab{}.
\newblock \showarticletitle{Unicoder-VL: A Universal Encoder for Vision and
  Language by Cross-modal Pre-training}.
\newblock \bibinfo{journal}{\emph{arXiv:1908.06066}} (\bibinfo{year}{2019}).
\newblock


\bibitem[\protect\citeauthoryear{Lin, Han, Mao, Wang, and Dally}{Lin
  et~al\mbox{.}}{2017}]%
        {lin2017deep}
\bibfield{author}{\bibinfo{person}{Yujun Lin}, \bibinfo{person}{Song Han},
  \bibinfo{person}{Huizi Mao}, \bibinfo{person}{Yu Wang}, {and}
  \bibinfo{person}{William~J Dally}.} \bibinfo{year}{2017}\natexlab{}.
\newblock \showarticletitle{Deep gradient compression: Reducing the
  communication bandwidth for distributed training}.
\newblock \bibinfo{journal}{\emph{arXiv preprint arXiv:1712.01887}}
  (\bibinfo{year}{2017}).
\newblock


\bibitem[\protect\citeauthoryear{Liu, Qiu, and Huang}{Liu
  et~al\mbox{.}}{2017}]%
        {liu2017adversarial}
\bibfield{author}{\bibinfo{person}{Pengfei Liu}, \bibinfo{person}{Xipeng Qiu},
  {and} \bibinfo{person}{Xuanjing Huang}.} \bibinfo{year}{2017}\natexlab{}.
\newblock \showarticletitle{Adversarial Multi-task Learning for Text
  Classification}. In \bibinfo{booktitle}{\emph{ACL}}. \bibinfo{pages}{1--10}.
\newblock


\bibitem[\protect\citeauthoryear{Liu, Ott, Goyal, Du, Joshi, Chen, Levy, Lewis,
  Zettlemoyer, and Stoyanov}{Liu et~al\mbox{.}}{2019}]%
        {liu2019roberta}
\bibfield{author}{\bibinfo{person}{Yinhan Liu}, \bibinfo{person}{Myle Ott},
  \bibinfo{person}{Naman Goyal}, \bibinfo{person}{Jingfei Du},
  \bibinfo{person}{Mandar Joshi}, \bibinfo{person}{Danqi Chen},
  \bibinfo{person}{Omer Levy}, \bibinfo{person}{Mike Lewis},
  \bibinfo{person}{Luke Zettlemoyer}, {and} \bibinfo{person}{Veselin
  Stoyanov}.} \bibinfo{year}{2019}\natexlab{}.
\newblock \showarticletitle{RoBERTa: A Robustly Optimized BERT Pretraining
  Approach}.
\newblock \bibinfo{journal}{\emph{arXiv:1907.11692}} (\bibinfo{year}{2019}).
\newblock


\bibitem[\protect\citeauthoryear{Lu, Batra, Parikh, and Lee}{Lu
  et~al\mbox{.}}{2019}]%
        {lu2019vilbert}
\bibfield{author}{\bibinfo{person}{Jiasen Lu}, \bibinfo{person}{Dhruv Batra},
  \bibinfo{person}{Devi Parikh}, {and} \bibinfo{person}{Stefan Lee}.}
  \bibinfo{year}{2019}\natexlab{}.
\newblock \showarticletitle{ViLBERT: Pretraining Task-Agnostic Visiolinguistic
  Representations for Vision-and-Language Tasks}.
\newblock \bibinfo{journal}{\emph{arXiv:1908.02265}} (\bibinfo{year}{2019}).
\newblock


\bibitem[\protect\citeauthoryear{Mou, Men, Li, Xu, Zhang, Yan, and Jin}{Mou
  et~al\mbox{.}}{2016a}]%
        {mou:acl2016}
\bibfield{author}{\bibinfo{person}{Lili Mou}, \bibinfo{person}{Rui Men},
  \bibinfo{person}{Ge Li}, \bibinfo{person}{Yan Xu}, \bibinfo{person}{Lu
  Zhang}, \bibinfo{person}{Rui Yan}, {and} \bibinfo{person}{Zhi Jin}.}
  \bibinfo{year}{2016}\natexlab{a}.
\newblock \showarticletitle{Natural Language Inference by Tree-Based
  Convolution and Heuristic Matching}. In \bibinfo{booktitle}{\emph{ACL}}.
\newblock


\bibitem[\protect\citeauthoryear{Mou, Meng, Yan, Li, Xu, Zhang, and Jin}{Mou
  et~al\mbox{.}}{2016b}]%
        {mou:EMNLP2016}
\bibfield{author}{\bibinfo{person}{Lili Mou}, \bibinfo{person}{Zhao Meng},
  \bibinfo{person}{Rui Yan}, \bibinfo{person}{Ge Li}, \bibinfo{person}{Yan Xu},
  \bibinfo{person}{Lu Zhang}, {and} \bibinfo{person}{Zhi Jin}.}
  \bibinfo{year}{2016}\natexlab{b}.
\newblock \showarticletitle{How Transferable are Neural Networks in NLP
  Applications?}. In \bibinfo{booktitle}{\emph{EMNLP}}.
\newblock


\bibitem[\protect\citeauthoryear{Pan, Wang, Qiu, Zhang, Li, and Huang}{Pan
  et~al\mbox{.}}{2021}]%
        {DBLP:conf/acl/Pan0QZLH20}
\bibfield{author}{\bibinfo{person}{Haojie Pan}, \bibinfo{person}{Chengyu Wang},
  \bibinfo{person}{Minghui Qiu}, \bibinfo{person}{Yichang Zhang},
  \bibinfo{person}{Yaliang Li}, {and} \bibinfo{person}{Jun Huang}.}
  \bibinfo{year}{2021}\natexlab{}.
\newblock \showarticletitle{Meta-KD: {A} Meta Knowledge Distillation Framework
  for Language Model Compression across Domains}. In
  \bibinfo{booktitle}{\emph{ACL/IJCNLP}}. \bibinfo{pages}{3026--3036}.
\newblock


\bibitem[\protect\citeauthoryear{Pan and Yang}{Pan and Yang}{2010}]%
        {pan2009survey}
\bibfield{author}{\bibinfo{person}{Sinno~Jialin Pan} {and}
  \bibinfo{person}{Qiang Yang}.} \bibinfo{year}{2010}\natexlab{}.
\newblock \showarticletitle{A Survey on Transfer Learning}.
\newblock \bibinfo{journal}{\emph{{IEEE} Trans. Knowl. Data Eng.}}
  \bibinfo{volume}{22}, \bibinfo{number}{10} (\bibinfo{year}{2010}),
  \bibinfo{pages}{1345--1359}.
\newblock


\bibitem[\protect\citeauthoryear{Peng}{Peng}{2020}]%
        {DBLP:journals/corr/abs-2004-11149}
\bibfield{author}{\bibinfo{person}{Huimin Peng}.}
  \bibinfo{year}{2020}\natexlab{}.
\newblock \showarticletitle{A Comprehensive Overview and Survey of Recent
  Advances in Meta-Learning}.
\newblock \bibinfo{journal}{\emph{arXiv:2004.11149}} (\bibinfo{year}{2020}).
\newblock


\bibitem[\protect\citeauthoryear{Qiu, Ma, Levy, tau Yih, Wang, and Tang}{Qiu
  et~al\mbox{.}}{2020}]%
        {qiu2020blockwise}
\bibfield{author}{\bibinfo{person}{Jiezhong Qiu}, \bibinfo{person}{Hao Ma},
  \bibinfo{person}{Omer Levy}, \bibinfo{person}{Scott~Wen tau Yih},
  \bibinfo{person}{Sinong Wang}, {and} \bibinfo{person}{Jie Tang}.}
  \bibinfo{year}{2020}\natexlab{}.
\newblock \showarticletitle{Blockwise Self-Attention for Long Document
  Understanding}.
\newblock \bibinfo{journal}{\emph{arXiv:1911.02972}} (\bibinfo{year}{2020}).
\newblock


\bibitem[\protect\citeauthoryear{Qu, Ji, Qiu, Yang, Min, Chen, Huang, and
  Croft}{Qu et~al\mbox{.}}{2019}]%
        {qu2019learning}
\bibfield{author}{\bibinfo{person}{Chen Qu}, \bibinfo{person}{Feng Ji},
  \bibinfo{person}{Minghui Qiu}, \bibinfo{person}{Liu Yang},
  \bibinfo{person}{Zhiyu Min}, \bibinfo{person}{Haiqing Chen},
  \bibinfo{person}{Jun Huang}, {and} \bibinfo{person}{W~Bruce Croft}.}
  \bibinfo{year}{2019}\natexlab{}.
\newblock \showarticletitle{Learning to selectively transfer: Reinforced
  transfer learning for deep text matching}. In
  \bibinfo{booktitle}{\emph{WSDM}}. \bibinfo{pages}{699--707}.
\newblock


\bibitem[\protect\citeauthoryear{Radford, Narasimhan, Salimans, and
  Sutskever}{Radford et~al\mbox{.}}{2018}]%
        {radford2018improving}
\bibfield{author}{\bibinfo{person}{Alec Radford}, \bibinfo{person}{Karthik
  Narasimhan}, \bibinfo{person}{Tim Salimans}, {and} \bibinfo{person}{Ilya
  Sutskever}.} \bibinfo{year}{2018}\natexlab{}.
\newblock \showarticletitle{Improving language understanding by generative
  pre-training}.
\newblock \bibinfo{journal}{\emph{arXiv}} (\bibinfo{year}{2018}).
\newblock


\bibitem[\protect\citeauthoryear{Radford, Wu, Child, Luan, Amodei, and
  Sutskever}{Radford et~al\mbox{.}}{2019}]%
        {radford2019language}
\bibfield{author}{\bibinfo{person}{Alec Radford}, \bibinfo{person}{Jeffrey Wu},
  \bibinfo{person}{Rewon Child}, \bibinfo{person}{David Luan},
  \bibinfo{person}{Dario Amodei}, {and} \bibinfo{person}{Ilya Sutskever}.}
  \bibinfo{year}{2019}\natexlab{}.
\newblock \showarticletitle{Language models are unsupervised multitask
  learners}.
\newblock \bibinfo{journal}{\emph{OpenAI blog}} \bibinfo{volume}{1},
  \bibinfo{number}{8} (\bibinfo{year}{2019}), \bibinfo{pages}{9}.
\newblock


\bibitem[\protect\citeauthoryear{Raffel, Shazeer, Roberts, Lee, Narang, Matena,
  Zhou, Li, and Liu}{Raffel et~al\mbox{.}}{2020}]%
        {raffel2020exploring}
\bibfield{author}{\bibinfo{person}{Colin Raffel}, \bibinfo{person}{Noam
  Shazeer}, \bibinfo{person}{Adam Roberts}, \bibinfo{person}{Katherine Lee},
  \bibinfo{person}{Sharan Narang}, \bibinfo{person}{Michael Matena},
  \bibinfo{person}{Yanqi Zhou}, \bibinfo{person}{Wei Li}, {and}
  \bibinfo{person}{Peter~J. Liu}.} \bibinfo{year}{2020}\natexlab{}.
\newblock \showarticletitle{Exploring the Limits of Transfer Learning with a
  Unified Text-to-Text Transformer}.
\newblock \bibinfo{journal}{\emph{J. Mach. Learn. Res.}}  \bibinfo{volume}{21}
  (\bibinfo{year}{2020}), \bibinfo{pages}{140:1--140:67}.
\newblock


\bibitem[\protect\citeauthoryear{Romero, Ballas, Kahou, Chassang, Gatta, and
  Bengio}{Romero et~al\mbox{.}}{2015}]%
        {DBLP:journals/corr/RomeroBKCGB14}
\bibfield{author}{\bibinfo{person}{Adriana Romero}, \bibinfo{person}{Nicolas
  Ballas}, \bibinfo{person}{Samira~Ebrahimi Kahou}, \bibinfo{person}{Antoine
  Chassang}, \bibinfo{person}{Carlo Gatta}, {and} \bibinfo{person}{Yoshua
  Bengio}.} \bibinfo{year}{2015}\natexlab{}.
\newblock \showarticletitle{FitNets: Hints for Thin Deep Nets}. In
  \bibinfo{booktitle}{\emph{ICLR}}.
\newblock


\bibitem[\protect\citeauthoryear{Ruder and Plank}{Ruder and Plank}{2017}]%
        {ruder2017learning}
\bibfield{author}{\bibinfo{person}{Sebastian Ruder} {and}
  \bibinfo{person}{Barbara Plank}.} \bibinfo{year}{2017}\natexlab{}.
\newblock \showarticletitle{Learning to select data for transfer learning with
  Bayesian Optimization}. In \bibinfo{booktitle}{\emph{EMNLP}}.
  \bibinfo{pages}{372--382}.
\newblock


\bibitem[\protect\citeauthoryear{Sanh, Debut, Chaumond, and Wolf}{Sanh
  et~al\mbox{.}}{2019}]%
        {DBLP:journals/corr/abs-1910-01108}
\bibfield{author}{\bibinfo{person}{Victor Sanh}, \bibinfo{person}{Lysandre
  Debut}, \bibinfo{person}{Julien Chaumond}, {and} \bibinfo{person}{Thomas
  Wolf}.} \bibinfo{year}{2019}\natexlab{}.
\newblock \showarticletitle{DistilBERT, a distilled version of {BERT:} smaller,
  faster, cheaper and lighter}.
\newblock \bibinfo{journal}{\emph{arXiv: 1910.01108}} (\bibinfo{year}{2019}).
\newblock


\bibitem[\protect\citeauthoryear{Schnabel and Sch{\"{u}}tze}{Schnabel and
  Sch{\"{u}}tze}{2014}]%
        {TSHS}
\bibfield{author}{\bibinfo{person}{Tobias Schnabel} {and}
  \bibinfo{person}{Hinrich Sch{\"{u}}tze}.} \bibinfo{year}{2014}\natexlab{}.
\newblock \showarticletitle{{FLORS:} Fast and Simple Domain Adaptation for
  Part-of-Speech Tagging}.
\newblock \bibinfo{journal}{\emph{Trans. Assoc. Comput. Linguistics}}
  \bibinfo{volume}{2} (\bibinfo{year}{2014}), \bibinfo{pages}{15--26}.
\newblock


\bibitem[\protect\citeauthoryear{Shazeer, Mirhoseini, Maziarz, Davis, Le,
  Hinton, and Dean}{Shazeer et~al\mbox{.}}{2017}]%
        {shazeer2017outrageously}
\bibfield{author}{\bibinfo{person}{Noam Shazeer}, \bibinfo{person}{Azalia
  Mirhoseini}, \bibinfo{person}{Krzysztof Maziarz}, \bibinfo{person}{Andy
  Davis}, \bibinfo{person}{Quoc Le}, \bibinfo{person}{Geoffrey Hinton}, {and}
  \bibinfo{person}{Jeff Dean}.} \bibinfo{year}{2017}\natexlab{}.
\newblock \bibinfo{title}{Outrageously Large Neural Networks: The
  Sparsely-Gated Mixture-of-Experts Layer}.
\newblock
\newblock
\showeprint[arxiv]{1701.06538}~[cs.LG]


\bibitem[\protect\citeauthoryear{Shen, Qu, Zhang, and Yu}{Shen
  et~al\mbox{.}}{2018}]%
        {shen2017wasserstein}
\bibfield{author}{\bibinfo{person}{Jian Shen}, \bibinfo{person}{Yanru Qu},
  \bibinfo{person}{Weinan Zhang}, {and} \bibinfo{person}{Yong Yu}.}
  \bibinfo{year}{2018}\natexlab{}.
\newblock \showarticletitle{Wasserstein Distance Guided Representation Learning
  for Domain Adaptation}. In \bibinfo{booktitle}{\emph{AAAI}}.
  \bibinfo{pages}{4058--4065}.
\newblock


\bibitem[\protect\citeauthoryear{Su, Zhu, Cao, Li, Lu, Wei, and Dai}{Su
  et~al\mbox{.}}{2020}]%
        {su2020vlbert}
\bibfield{author}{\bibinfo{person}{Weijie Su}, \bibinfo{person}{Xizhou Zhu},
  \bibinfo{person}{Yue Cao}, \bibinfo{person}{Bin Li}, \bibinfo{person}{Lewei
  Lu}, \bibinfo{person}{Furu Wei}, {and} \bibinfo{person}{Jifeng Dai}.}
  \bibinfo{year}{2020}\natexlab{}.
\newblock \showarticletitle{VL-BERT: Pre-training of Generic Visual-Linguistic
  Representations}.
\newblock \bibinfo{journal}{\emph{arXiv:1908.08530}} (\bibinfo{year}{2020}).
\newblock


\bibitem[\protect\citeauthoryear{Sun, Cheng, Gan, and Liu}{Sun
  et~al\mbox{.}}{2019}]%
        {DBLP:journals/corr/abs-1908-09355}
\bibfield{author}{\bibinfo{person}{Siqi Sun}, \bibinfo{person}{Yu Cheng},
  \bibinfo{person}{Zhe Gan}, {and} \bibinfo{person}{Jingjing Liu}.}
  \bibinfo{year}{2019}\natexlab{}.
\newblock \showarticletitle{Patient Knowledge Distillation for {BERT} Model
  Compression}. In \bibinfo{booktitle}{\emph{EMNLP-IJCNLP}}.
  \bibinfo{pages}{4322--4331}.
\newblock


\bibitem[\protect\citeauthoryear{Tan, Sun, Kong, Zhang, Yang, and Liu}{Tan
  et~al\mbox{.}}{2018}]%
        {tan2018survey}
\bibfield{author}{\bibinfo{person}{Chuanqi Tan}, \bibinfo{person}{Fuchun Sun},
  \bibinfo{person}{Tao Kong}, \bibinfo{person}{Wenchang Zhang},
  \bibinfo{person}{Chao Yang}, {and} \bibinfo{person}{Chunfang Liu}.}
  \bibinfo{year}{2018}\natexlab{}.
\newblock \showarticletitle{A Survey on Deep Transfer Learning}. In
  \bibinfo{booktitle}{\emph{ICANN}}. \bibinfo{pages}{270--279}.
\newblock


\bibitem[\protect\citeauthoryear{Tang, Lu, Liu, Mou, Vechtomova, and Lin}{Tang
  et~al\mbox{.}}{2019}]%
        {DBLP:journals/corr/abs-1903-12136}
\bibfield{author}{\bibinfo{person}{Raphael Tang}, \bibinfo{person}{Yao Lu},
  \bibinfo{person}{Linqing Liu}, \bibinfo{person}{Lili Mou},
  \bibinfo{person}{Olga Vechtomova}, {and} \bibinfo{person}{Jimmy Lin}.}
  \bibinfo{year}{2019}\natexlab{}.
\newblock \showarticletitle{Distilling Task-Specific Knowledge from {BERT} into
  Simple Neural Networks}.
\newblock \bibinfo{journal}{\emph{arXiv: 1903.12136}} (\bibinfo{year}{2019}).
\newblock


\bibitem[\protect\citeauthoryear{Vaswani, Shazeer, Parmar, Uszkoreit, Jones,
  Gomez, Kaiser, and Polosukhin}{Vaswani et~al\mbox{.}}{2017}]%
        {vaswani2017attention}
\bibfield{author}{\bibinfo{person}{Ashish Vaswani}, \bibinfo{person}{Noam
  Shazeer}, \bibinfo{person}{Niki Parmar}, \bibinfo{person}{Jakob Uszkoreit},
  \bibinfo{person}{Llion Jones}, \bibinfo{person}{Aidan~N. Gomez},
  \bibinfo{person}{Lukasz Kaiser}, {and} \bibinfo{person}{Illia Polosukhin}.}
  \bibinfo{year}{2017}\natexlab{}.
\newblock \showarticletitle{Attention is All you Need}. In
  \bibinfo{booktitle}{\emph{NIPS}}. \bibinfo{pages}{5998--6008}.
\newblock


\bibitem[\protect\citeauthoryear{Wang, Qiu, Wang, Li, Gong, Zeng, Huang, Zheng,
  Cai, and Zhou}{Wang et~al\mbox{.}}{2019}]%
        {mgtl}
\bibfield{author}{\bibinfo{person}{B. Wang}, \bibinfo{person}{M. Qiu},
  \bibinfo{person}{X. Wang}, \bibinfo{person}{Y. Li}, \bibinfo{person}{Y.
  Gong}, \bibinfo{person}{X. Zeng}, \bibinfo{person}{J. Huang},
  \bibinfo{person}{B. Zheng}, \bibinfo{person}{D. Cai}, {and}
  \bibinfo{person}{J. Zhou}.} \bibinfo{year}{2019}\natexlab{}.
\newblock \showarticletitle{A Minimax Game for Instance based Selective
  Transfer Learning}. In \bibinfo{booktitle}{\emph{SIGKDD}}.
\newblock


\bibitem[\protect\citeauthoryear{Wang, Qiu, Huang, and He}{Wang
  et~al\mbox{.}}{2020b}]%
        {emnlp2020mft}
\bibfield{author}{\bibinfo{person}{C. Wang}, \bibinfo{person}{M. Qiu},
  \bibinfo{person}{J. Huang}, {and} \bibinfo{person}{X. He}.}
  \bibinfo{year}{2020}\natexlab{b}.
\newblock \showarticletitle{Meta Fine-Tuning Neural Language Models for
  Multi-Domain Text Mining}. In \bibinfo{booktitle}{\emph{EMNLP}}.
\newblock


\bibitem[\protect\citeauthoryear{Wang, Li, Khabsa, Fang, and Ma}{Wang
  et~al\mbox{.}}{2020a}]%
        {wang2020linformer}
\bibfield{author}{\bibinfo{person}{Sinong Wang}, \bibinfo{person}{Belinda~Z.
  Li}, \bibinfo{person}{Madian Khabsa}, \bibinfo{person}{Han Fang}, {and}
  \bibinfo{person}{Hao Ma}.} \bibinfo{year}{2020}\natexlab{a}.
\newblock \showarticletitle{Linformer: Self-Attention with Linear Complexity}.
\newblock \bibinfo{journal}{\emph{arXiv:2006.04768}} (\bibinfo{year}{2020}).
\newblock


\bibitem[\protect\citeauthoryear{Williams, Nangia, and Bowman}{Williams
  et~al\mbox{.}}{2018}]%
        {MultiNLI}
\bibfield{author}{\bibinfo{person}{Adina Williams}, \bibinfo{person}{Nikita
  Nangia}, {and} \bibinfo{person}{Samuel~R. Bowman}.}
  \bibinfo{year}{2018}\natexlab{}.
\newblock \showarticletitle{A Broad-Coverage Challenge Corpus for Sentence
  Understanding through Inference}. In \bibinfo{booktitle}{\emph{NAACL}}.
\newblock


\bibitem[\protect\citeauthoryear{Yang, Salakhutdinov, and Cohen.}{Yang
  et~al\mbox{.}}{2017}]%
        {Zhilin}
\bibfield{author}{\bibinfo{person}{Zhilin Yang}, \bibinfo{person}{Ruslan
  Salakhutdinov}, {and} \bibinfo{person}{William~W. Cohen.}}
  \bibinfo{year}{2017}\natexlab{}.
\newblock \showarticletitle{Transfer Learning for Sequence Tagging with
  Hierarchical Recurrent Networks}. In \bibinfo{booktitle}{\emph{ICLR}}.
\newblock


\bibitem[\protect\citeauthoryear{Yu, Qiu, Jiang, Huang, Song, Chu, and Chen}{Yu
  et~al\mbox{.}}{2018}]%
        {drss}
\bibfield{author}{\bibinfo{person}{Jianfei Yu}, \bibinfo{person}{Minghui Qiu},
  \bibinfo{person}{Jing Jiang}, \bibinfo{person}{Jun Huang},
  \bibinfo{person}{Shuangyong Song}, \bibinfo{person}{Wei Chu}, {and}
  \bibinfo{person}{Haiqing Chen}.} \bibinfo{year}{2018}\natexlab{}.
\newblock \showarticletitle{Modelling Domain Relationships for Transfer
  Learning on Retrieval-based Question Answering Systems in E-commerce}. In
  \bibinfo{booktitle}{\emph{WSDM}}.
\newblock


\bibitem[\protect\citeauthoryear{Zhang, Li, Tao, Yang, Tang, and Xu}{Zhang
  et~al\mbox{.}}{2014}]%
        {fuxi}
\bibfield{author}{\bibinfo{person}{Zhuo Zhang}, \bibinfo{person}{Chao Li},
  \bibinfo{person}{Yangyu Tao}, \bibinfo{person}{Renyu Yang},
  \bibinfo{person}{Hong Tang}, {and} \bibinfo{person}{Jie Xu}.}
  \bibinfo{year}{2014}\natexlab{}.
\newblock \showarticletitle{Fuxi: a Fault-Tolerant Resource Management and Job
  Scheduling System at Internet Scale}.
\newblock \bibinfo{journal}{\emph{Proc. {VLDB} Endow.}} \bibinfo{volume}{7},
  \bibinfo{number}{13} (\bibinfo{year}{2014}), \bibinfo{pages}{1393--1404}.
\newblock


\bibitem[\protect\citeauthoryear{Zheng, Zhao, Long, Zhu, Zhu, Zhao, Diao, Yang,
  and Lin}{Zheng et~al\mbox{.}}{2020}]%
        {zheng2020fusionstitching}
\bibfield{author}{\bibinfo{person}{Zhen Zheng}, \bibinfo{person}{Pengzhan
  Zhao}, \bibinfo{person}{Guoping Long}, \bibinfo{person}{Feiwen Zhu},
  \bibinfo{person}{Kai Zhu}, \bibinfo{person}{Wenyi Zhao},
  \bibinfo{person}{Lansong Diao}, \bibinfo{person}{Jun Yang}, {and}
  \bibinfo{person}{Wei Lin}.} \bibinfo{year}{2020}\natexlab{}.
\newblock \showarticletitle{Fusionstitching: boosting memory intensive
  computations for deep learning workloads}.
\newblock \bibinfo{journal}{\emph{arXiv:2009.10924}} (\bibinfo{year}{2020}).
\newblock


\bibitem[\protect\citeauthoryear{Zhuang, Huang, He, Ma, and He}{Zhuang
  et~al\mbox{.}}{2017}]%
        {TLMRCNN}
\bibfield{author}{\bibinfo{person}{Fuzhen Zhuang}, \bibinfo{person}{Lang
  Huang}, \bibinfo{person}{Jia He}, \bibinfo{person}{Jixin Ma}, {and}
  \bibinfo{person}{Qing He}.} \bibinfo{year}{2017}\natexlab{}.
\newblock \showarticletitle{Transfer Learning with Manifold Regularized
  Convolutional Neural Network}. In \bibinfo{booktitle}{\emph{KSEM}}.
\newblock


\bibitem[\protect\citeauthoryear{Zhuang, Qi, Duan, Xi, Zhu, Zhu, Xiong, and
  He}{Zhuang et~al\mbox{.}}{2019}]%
        {DBLP:journals/corr/abs-1911-02685}
\bibfield{author}{\bibinfo{person}{Fuzhen Zhuang}, \bibinfo{person}{Zhiyuan
  Qi}, \bibinfo{person}{Keyu Duan}, \bibinfo{person}{Dongbo Xi},
  \bibinfo{person}{Yongchun Zhu}, \bibinfo{person}{Hengshu Zhu},
  \bibinfo{person}{Hui Xiong}, {and} \bibinfo{person}{Qing He}.}
  \bibinfo{year}{2019}\natexlab{}.
\newblock \showarticletitle{A Comprehensive Survey on Transfer Learning}.
\newblock \bibinfo{journal}{\emph{arXiv:1911.02685}} (\bibinfo{year}{2019}).
\newblock


\end{thebibliography}

\end{document}